\documentclass[conference]{IEEEtran}
\usepackage{times}
\usepackage{url}
\usepackage{amsfonts}
\usepackage{amsmath}
\usepackage{stfloats}
\usepackage{float}
\usepackage{algorithm}  
\usepackage{algorithmicx}  
\usepackage{algpseudocode}  
\usepackage{amsmath}  
\usepackage[T1]{fontenc}
\floatname{algorithm}{Algorithm}

\usepackage[numbers]{natbib}
\usepackage{multicol}
\usepackage[bookmarks=true]{hyperref}
\usepackage{graphicx}
\usepackage{authblk}

\begin{document}

\title{Fault Tolerant Free Gait and Footstep Planning for Hexapod Robot Based on Monte-Carlo Tree}


\author[1,*]{Liang Ding} 
\author[1,*]{Peng Xu}
\author[1]{Haibo Gao}
\author[1]{Zhikai Wang}
\author[1]{Ruyi Zhou}
\author[1]{Zhaopei Gong}
\author[2]{Guangjun Liu}
\affil[1]{State
Key Laboratory of Robotics and Systems, Harbin Institute of Technology, Harbin 150001, China}
\affil[2]{ Department of Aerospace
Engineering, Ryerson University, Toronto, ON M5B 2K3, Canada}
\affil[*]{These authors contributed equally to this work. Contact Email: {liangding@hit.edu.cn}}




\maketitle



\begin{abstract}
Legged robots can pass through complex field environments by selecting gaits and discrete footholds carefully. Traditional methods plan gait and foothold separately and treat them as the single-step optimal process. However, such processing causes its poor passability in a sparse foothold environment. This paper novelly proposes a coordinative planning method for hexapod robots that regards the planning of gait and foothold as a sequence optimization problem with the consideration of dealing with the harshness of the environment as leg fault. The Monte Carlo tree search algorithm(MCTS) is used to optimize the entire sequence. Two methods, FastMCTS, and SlidingMCTS are proposed to solve some defeats of the standard MCTS applicating in the field of legged robot planning. The proposed planning algorithm combines the fault-tolerant gait method to improve the passability of the algorithm. Finally, compared with other planning methods, experiments on terrains with different densities of footholds and artificially-designed challenging terrain are carried out to verify our methods. All results show that the proposed method dramatically improves the hexapod robot’s ability to pass through sparse footholds environment.
\end{abstract}

\IEEEpeerreviewmaketitle

\section{Introduction}

Legged robots can select discrete footholds to cross various complicated terrain, which leads them to execute motor tasks on fields such as field rescue and planetary exploration in the future. The hexapod robots that have higher stability and superior load capacity than biped robots and quadruped robots are widely used\cite{moore2002reliable}\cite{tunc2016experimental}\cite{picardi2020bioinspired}. Still, the planning of gait and foothold for such robots is more complicated with more legs. The traditional planning framework plans gait first, and then plan the foothold of the swing leg according to the terrain\cite{kalakrishnan2011learning}\cite{belter2016adaptive}\cite{fankhauser2018robust}. The two steps are independent of each other, making optimal decisions based on corresponding rules or evaluation functions. However, in harsh environments, traditional planning frameworks can easily cause robots to be trapped because such planning methods make decisions are based on only the current environment and the state of the robot, and it does not consider the following situation. In this paper, We focus on improving the robot’s ability to pass in a sparse foothold environment by selecting appropriate footholds and gait.

 Gait is usually used to express the walking mode of the legged robot. The choice of gait can affect the robot’s forward speed, stability, and passability. Classifying by whether the gait changes periodically, there are two modes, including periodic gait and aperiodic gait. According to different planning methods, gait can be divided into the rule-based method and CPG method. For rule-based method, when walking in a periodic gait, assuming that all footsteps are valid, legged robots move forward in a fixed swing sequence, which is usually taken as 3+3 tripod gait, 4+2 quadruped gait or 5+1 wave gait for hexapod robots\cite{chu2002comparison}. Because these gaits are quickly to use, they are widely used by researchers\cite{estremera2010continuous}\cite{bjelonic2018weaver}\cite{belter2019employing}. When the terrain is rugged or some areas are unsupportable, the legged robot needs to change its gait according to the terrain information and its state information, then generate a sequence of gaits with an irregular order. This kind of aperiodic gait is called as free gait. The free gait proposed by Kugushev and Jaroshevskij\cite{kugushev1975problems}  in 1975 is characterized by aperiodic, irregular, asymmetric, and terrain adapt. For free gait, the order of legs changes in a non-fixed but flexible manner depending on the trajectory, terrain properties and motion state. In irregular terrain, this gait type is more flexible and adaptable than periodic and regular gait. A large number of free gaits for quadruped or hexapod robots have been developed so far\cite{mcghee1979adaptive}\cite{hirose1984study}\cite{estremera2005generating}.\par

 \begin{figure}[tbp] 
\centering 
\includegraphics[scale=0.13]{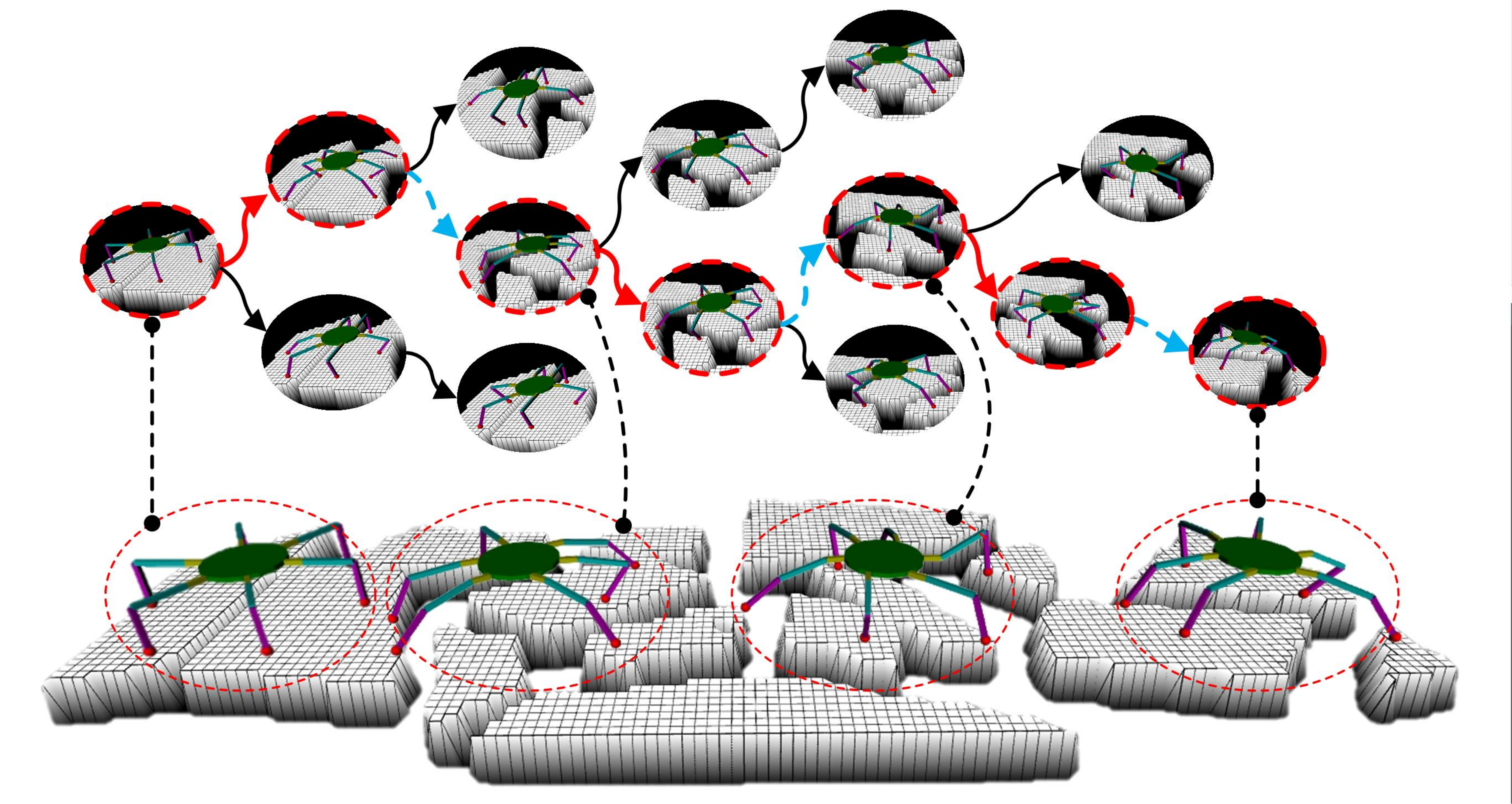} 
\caption{Establishing Monte Carlo tree for walking planning of hexapod robot in sparse foothold environment. Each node represents a state of the robot. When the constructed node reaches the target position, the entire search algorithm ends. The dotted arrows in the figure indicate omitted parts, and the whole tree in the figure is a schematic structure and is not complete.} 
\label{fig1} 
\end{figure}


Another biologically-inspired gait generation method is the CPG method. From the perspective of imitating the biological gait rhythm control, the CPG gait planning method regards each foot of the robot as a neuron and realizes the walking of the robot by periodically triggering the movement of each foot. Shik et al.\cite{shik1966control} proposed that the rhythmic movement of animals was controlled by a central pattern generator (CPG). Since then, a large number of scholars have carried out studies on CPG to planning the gait for legged robots  \cite{venkataraman1997simple}\cite{arena2004adaptive}\cite{liu2013central}. CPG gait mainly controls the movement of the legs through an oscillator, without any feedback, and it is easy to achieve smooth transitions between gait. However, when the environment becomes complicated, if a sequence of legs is wrong, the whole CPG model will collapse. Therefore some scholars combined it with the reflex model to improve the environmental adaptability of the CPG model. For example, Santos\cite{santos2011gait} used the reflex model to provide reflection signals and realized the dynamic regulation of the movement rhythm by modifying the CPG model parameters. Mustafa Suphi Erden\cite{erden2008free} used the reinforcement learning method to conduct the CPG network and reflection model's structural training.\par


The selection of foothold is often carried out after gait planning. For foothold planning method, Expert threshold method is commonly used by scholars to select the footholds according to the features such as the roughness of the terrain, the amount of slope, the degree of proximity to the edge, the amount of slip and the height variance \cite{krotkov1996perception}\cite{rebula2007controller}. Kolter\cite{kolter2008hierarchical}  used a hierarchical apprenticeship learning algorithm to select the footholds. It still used the human expert experience to adjust the cost function weights. Besides, Kalakrishnan\cite{kalakrishnan2009learning}  proposed a more elaborate method, using geometric terrain template learning to extract useful landing features, and the terrain template was completed by human teaching. \cite{belter2011rough}  established a 2.5D map through 2D lidar, and propose a foothold selection algorithm, which employs unsupervised learning to create an adaptive decision surface. The robot can learn from realistic simulations, and no prior expert-given rules or parameters are used. The above methods only consider the environmental characteristics where the robot is. When the environment becomes extremely harsh, if a leg has no foothold, no related work has been found to solve it. Our method is that the problem can be avoided by sequence optimization, or a fault-tolerant gait method can be combined to deal with it.\par


Inspired by well-known artificial intelligence case AlphaGo\cite{silver2016mastering}\cite{silver2017mastering}, Mente-Carlo Tree Search(MCTS)\cite{browne2012survey} is an excellent method to find an optimal decision to solve the sequence optimization problem. Monte Carlo methods have a long history within numerical algorithms and have also had significant success in various AI game-playing algorithms. Recently, Monte Carlo trees have been used in unmanned vehicles and robots. For example, \cite{lenz2016tactical} adopted the MCTS algorithm to consider interactions between different vehicles to plan cooperative motion plans.  \cite{naghshvar2018risk} combines QMDP, unscented transform, and MCTS to establish an autonomous driving decision framework. For the first time, MCTS was used to solve the planning problem of legged robots\cite{clary2018monte}. This work mainly demonstrates the application of the MCTS method to the field of blind walking of biped robots, which requires robots to avoid obstacles on the platform ground. \par


For other sequence optimization methods, there are several works.  \cite{aceituno2017simultaneous} combines contact search, and trajectory generation into a Mixed-Integer Convex Optimization problem at the same time, a sequence optimization method was formed.  \cite{naderi2017discovering} combines a graph-based high-level path planner with low-level sampling-based optimization of climbing to plan a footstep sequence. In the work of  \cite{tsounis2019deepgait}, the quadruped robot Anymal was trained to walk in complex environments through Deep Reinforcement Learning. They trained the perceptual planning layer and control layer into two networks. The perceptual planning layer strategy can generate basic motion sequences that lead the robot to the target position. The method is similar to the sequence optimization problem we emphasized, but they do not focus on the comparison of such passability with traditional methods. Whether the superiority in the sparse foothold environment can be guaranteed is not carefully explained. Besides, the above optimization work is carried out on quadruped or biped robots. Hexapod robots have a richer combination of gait and foothold, and no related work has been described yet.\par

In this article, we mainly discuss how to plan gait and foothold to improve the robot's ability to pass in sparse foothold. The main contributions of this paper lie in: \par

1) The gait generation and foothold planning are solved as a sequence optimization problem, and the Monte Carlo Tree Search algorithm is used to optimize the decision sequence. Method couples gait generation and foothold selection.\par

2) Treat the legs without candidate foothold as faulty legs, and combine the idea of fault-tolerant gait with our planning method to improve the passing ability of hexapod robots in extreme environments. In addition, a free fault-tolerant gait expert planning method considering environmental fault tolerance is also proposed.\par

3) Two methods, Fast-MCTS, and Sliding-MCTS are proposed. The Fast-MCTS method has higher pass performance and faster search speed. Sliding-MCTS has an effective balance between optimization and search time.\par

4) Compare the indicators of traditional methods and sequence optimization methods in the sparse foothold environment. The advantages and disadvantages of different methods are explained.

\section{Fault Tolerant Free Gait Planning}
To explain the method better, first define and explain the relevant indicators for gait and foothold planning of hexapod robot.

\subsection{Notation and Definition}

\begin{figure}[ht] 
\centering 
\includegraphics[scale=0.6]{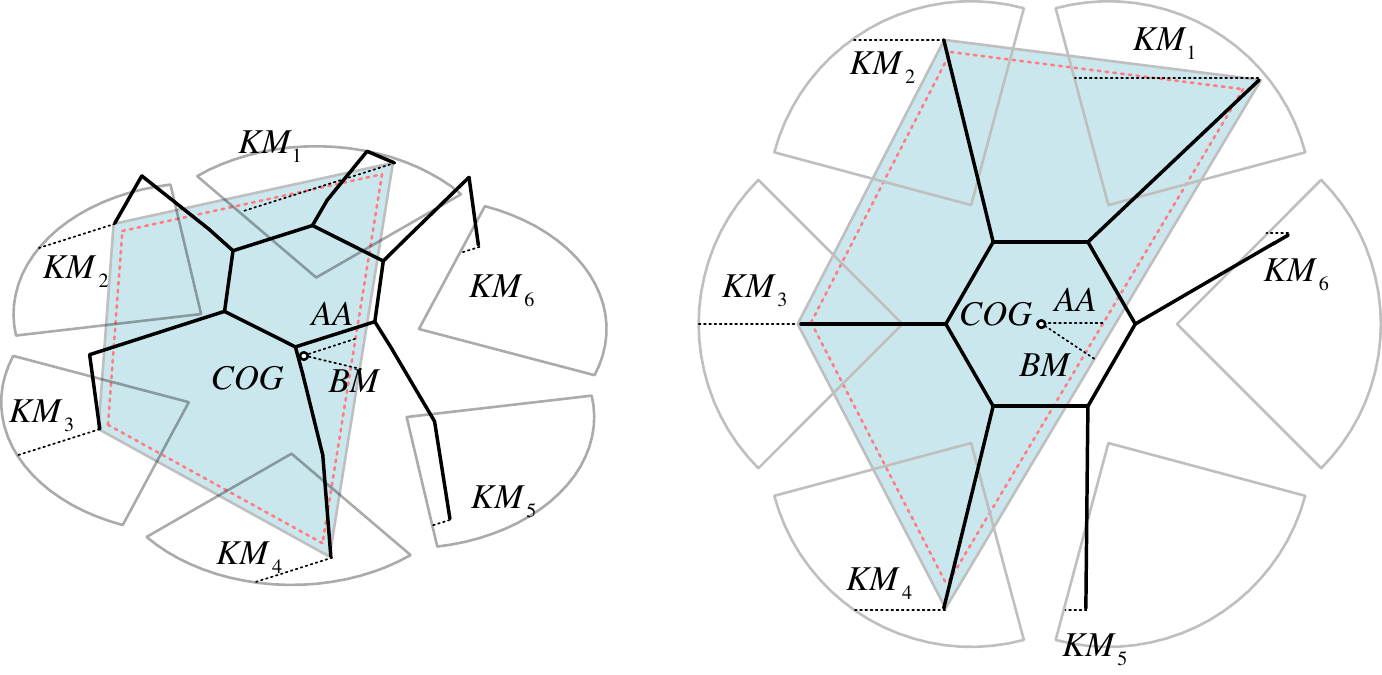} 
\caption{Parameter definition of hexapod robot planning } 
\label{hexapodAxis} 
\end{figure}

\noindent  \textbf{Definition}:Support polygon\par
\begin{figure}[ht] 
\centering 
\includegraphics[scale=0.9]{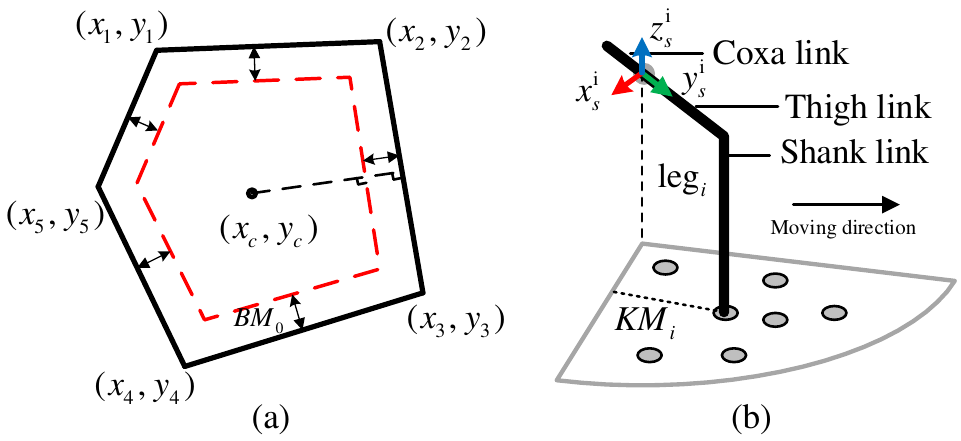} 
\caption{(a) Support polygons shrink in proportion. (b) Simplified one-leg workspace.} 
\label{scaledPolygon} 
\end{figure}

The support polygon is a convex polygon formed by the projection points of the robot's supporting feet positions falling on the horizontal plane. Support polygons are often used to measure the stability of legged robot. If the horizontal projection of robot's COG falls within the supporting polygon, then the robot is statically stable. When robot moves, if the center of gravity is too close to the edge of the supporting polygon, the stability of robot is poor. In order to reduce the critical stability process during the planning process, this paper uses the centroid as center to reduce the support polygon, As shown in Figure  \ref {scaledPolygon}(a), the polygon formed by the solid line is the original supporting polygon, and the polygon formed by the broken line is the reduced supporting polygon. $(x_c,y_c)$ represents the coordinate of centroid, and  $(x_i,y_i)$ denotes the coordinate of one support leg's feet position. The formula for calculating centroid coordinates $(x_c,y_c)$ is as follows:\par
\begin{equation} 
x_c = \frac{1}{6A}\sum_{i=1}^{n}(x_i+x_{i+1})(x_i\cdot y_{i+1} - x_{i+1}\cdot y_i )
\end{equation}
\begin{equation} 
y_c = \frac{1}{6A}\sum_{i=1}^{n}(y_i+y_{i+1})(x_i\cdot y_{i+1} - x_{i+1}\cdot y_i )
\end{equation}
Where A represents the area of the original supporting polygon.

\begin{equation} 
A = \frac{1}{2}\sum_{i=1}^{n}(x_i\cdot y_{i+1} - x_{i+1}\cdot y_i )
\end{equation}
Finally, according to the calculated centroid position and a constant stability margin $BM_0$, the support polygon is reduced. 

\noindent  \textbf{Definition}:Single Leg Workspace  \par
The workspace of a single leg is simplified into a fan-shaped space, as shown in Figure \ref{scaledPolygon}(b). The sector is defined in the single-leg workspace coordinate system $\sum_{s_i}(O_{s_i}-x_s^{(i)}y_s^{(i)}z_s^{(i)})$ , Coordinate system $\sum_{s_i}$ is fixed to the robot.\newline

\noindent  \textbf{Definition}:Support State\par
$c_f:= [s_1,s_2,s_3,s_4,s_5,s_6]\in \mathbb{R}^{1\times 6}$ is a vector indicating the support state of hexapod moving to the next step. If leg $ i $ is a support leg, the value of $ s_i $ is 0, if the leg $ i $ is a swing leg, the value of $ s_i $ is 1. \newline

\noindent  \textbf{Definition}:Fault Leg \par
If the environment is very complicated, which results in that some legs do not have effective footholds to choose from. Then at this time, the leg with no alternative foothold is defined as the fault leg. For possible physical damages to a leg, we also treat it as a fault leg.\newline\par

\noindent  \textbf{Definition}:Leg Fault State\par
$t_F:=[f_1,f_2,f_3,f_4,f_5,f_6]\in \mathbb{R}^{1\times 6}$ represents the leg fault state vector of the six-legged robot from the current state to the next state. If leg $i$ is a fault leg, the value of $f_i$ is 1. If leg $i$ is a normal leg, the value of $f_i$ is 0. Note that if a leg is the fault leg, it must not be a support leg.\newline\par

\noindent  \textbf{Definition}:Hexapod State   \par
$\Phi:= <$ $_B^W\!R,\  _{B}^{W}\!{r}\ , c_F,\ t_F,\  _{F}^{W}\textrm{r}$ $> $ is defined as the state of hexapod robot. Where $_B^W\!R\in SO_3$ is the rotation matrix representing the attitude of the base w.r.t $W$ frame. $_{B}^{W}\!{r}\in  \mathbb{R}^{3} $ is the target position of robot's COG in next step w.r.t $W$ frame. $c_F$ is the support state vector of the robot from the current state to the next state. $t_F$ represents the leg fault state vector of the hexapod robot from the current state to the next state. $_{F}^{W}\!{r}\in  \mathbb{R}^{3} $ is the target position of $i_{th}$ foot in next step(foothold position) w.r.t $W$ frame.
\newline

\noindent  \textbf{Definition}:Static Margin\par
Stability margin, $SM$, also known as the absolute static stability margin, is the smallest distance from the vertical projection of the COG(centre of gravity) on a horizontal plane to the sides of the support polygon formed by joining the projections of the footholds on the same horizontal plane, as is shown in Figure \ref{hexapodAxis}.\newline

\noindent  \textbf{Definition}:Reduced Kinematic Margin\par
As shown in Figure \ref{scaledPolygon}(b), the reduced kinematic margin,  $KM_i$, represents the distance that the $i_{th}$ foot position moves in the opposite direction of the robot motion and reaches the boundary of the working space of leg $i$. \newline\par

\noindent  \textbf{Definition}:Maximum Advance Amount Based COG\par
The maximum advance amount based support area is the maximum distance which the hexapod can moves in the forward direction in the condition that the COG can’t exceed the support area. It is defined as $AA$.\newline

\noindent  \textbf{Definition}:Maximum Step Length\par
The maximum step length is the maximum distance that hexapod can move as long as in the forward direction. It depends on the hexapod’s state and is defined as:
\begin{equation}  \label{MSL}
MSL={\rm min}(KM_i,AA)(i=1,2,3...6)
\end{equation}
\begin{figure*}[ht] 
\centering 
\includegraphics[scale=0.8]{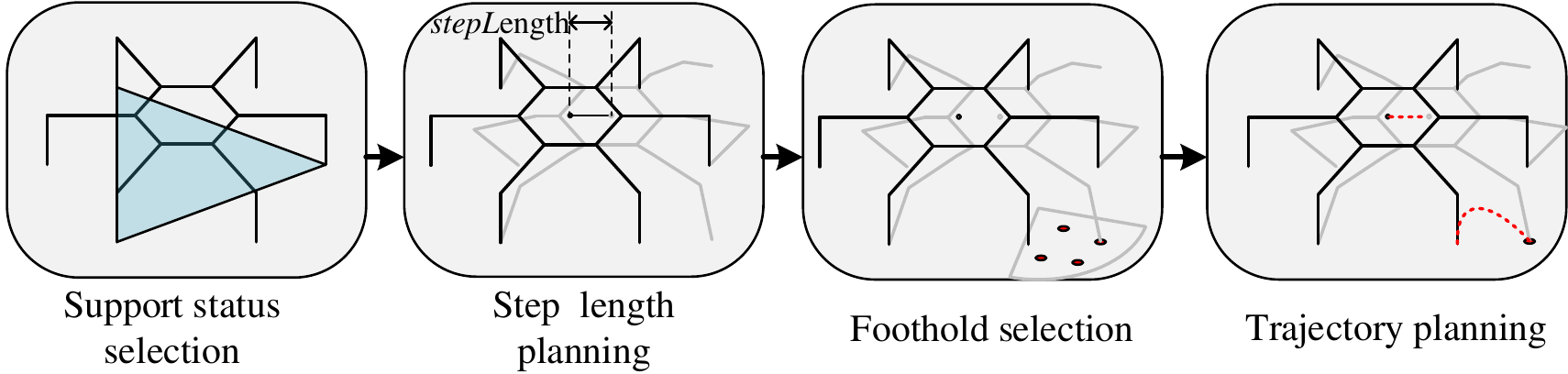} 
\caption{Traditional legged robot motion planning framework.} 
\label{planningPipline} 
\end{figure*}

\noindent  \textbf{Definition}:Support State List \par
Support State List represents the maximum allowable set of support states for the robot. Each leg of a legged robot has support state and swing state two states. The combination of different leg states constitutes the support state of the robot. For a hexapod robot, there are $2^6$, 64 possible support states. To ensure static stability, the number of supporting legs should not be less than 3. After excluding these support states, only 42 alternative support states remain, as shown in Table \ref{SupportStateList}.When planning the next support state with a specified six-footed robot state, the supportable state table is used as an initial candidate state table for screening, and a new candidate support state table that meets the requirements is finally obtained.\newline\par

\noindent  \textbf{Definition}:Solution Sequence \par
The solution sequence represents the state sequence of the robot from the current position to the target position. Define the solution sequence as $\Psi=\left \{ \Phi_1,\Phi_2...\Phi_k \right \} $, Indicates that the robot needs to go through $k$ state transitions to reach its destination.

\begin{table}[h]
\caption{Support State List}
\label{SupportStateList} 
\begin{tabular}{cccccccccccccc}
\hline      
\textbf{Num} & \multicolumn{6}{c}{ \textbf{Support State}} & \textbf{Num} & \multicolumn{6}{c}{\textbf{Support State}}\\
\hline      
1   & 0   & 0   & 0   & 1   & 1   & 1   & 22  & 1   & 0   & 1   & 0   & 1   & 0   \\
2   & 0   & 0   & 1   & 0   & 1   & 1   & 23  & 1   & 0   & 1   & 0   & 1   & 1   \\
3   & 0   & 0   & 1   & 1   & 0   & 1   & 24  & 1   & 0   & 1   & 1   & 0   & 0   \\
4   & 0   & 0   & 1   & 1   & 1   & 0   & 25  & 1   & 0   & 1   & 1   & 0   & 1   \\
5   & 0   & 0   & 1   & 1   & 1   & 1   & 26  & 1   & 0   & 1   & 1   & 1   & 0   \\
6   & 0   & 1   & 0   & 0   & 1   & 1   & 27  & 1   & 0   & 1   & 1   & 1   & 1   \\
7   & 0   & 1   & 0   & 1   & 0   & 1   & 28  & 1   & 1   & 0   & 0   & 0   & 1   \\
8   & 0   & 1   & 0   & 1   & 1   & 0   & 29  & 1   & 1   & 0   & 0   & 1   & 0   \\
9   & 0   & 1   & 0   & 1   & 1   & 1   & 30  & 1   & 1   & 0   & 0   & 1   & 1   \\
10  & 0   & 1   & 1   & 0   & 0   & 1   & 31  & 1   & 1   & 0   & 1   & 0   & 0   \\
11  & 0   & 1   & 1   & 0   & 1   & 0   & 32  & 1   & 1   & 0   & 1   & 0   & 1   \\
12  & 0   & 1   & 1   & 0   & 1   & 1   & 33  & 1   & 1   & 0   & 1   & 1   & 0   \\
13  & 0   & 1   & 1   & 1   & 0   & 0   & 34  & 1   & 1   & 0   & 1   & 1   & 1   \\
14  & 0   & 1   & 1   & 1   & 0   & 1   & 35  & 1   & 1   & 1   & 0   & 0   & 0   \\
15  & 0   & 1   & 1   & 1   & 1   & 0   & 36  & 1   & 1   & 1   & 0   & 0   & 1   \\
16  & 0   & 1   & 1   & 1   & 1   & 1   & 37  & 1   & 1   & 1   & 0   & 1   & 0   \\
17  & 1   & 0   & 0   & 0   & 1   & 1   & 38  & 1   & 1   & 1   & 0   & 1   & 1   \\
18  & 1   & 0   & 0   & 1   & 0   & 1   & 39  & 1   & 1   & 1   & 1   & 0   & 0   \\
19  & 1   & 0   & 0   & 1   & 1   & 0   & 40  & 1   & 1   & 1   & 1   & 0   & 1   \\
20  & 1   & 0   & 0   & 1   & 1   & 1   & 41  & 1   & 1   & 1   & 1   & 1   & 0   \\
21  & 1   & 0   & 1   & 0   & 0   & 1   & 42  & 1   & 1   & 1   & 1   & 1   & 1  \\
\hline
\end{tabular}
\end{table}

\subsection{Fault Tolerant Free Gait Method}

\noindent(1) Traditional free gait planning pipline\par
The free fault-tolerant gait proposed in this section is based on the traditional expert planning process. Traditional expert planning pipeline is shown in Figure \ref{planningPipline}. First, plan the support state $c_F$ (gait). Second, according to the selected gait, the step length of the robot is determined. Third, for the swing legs determined by step 1, the optimal footholds $_{F}^{W}\!{r}$  are selected in their working space. Finally, According to the terminal robot state $\Phi_{k+1}$, current robot state $\Phi_{k}$  and environmental obstacle information planned in the above steps, the robot body and leg trajectory are generated.\par

To improve the stability and passiblity, the traditional planning framework also contains posture optimization step. This article focuses on gait and foothold planning, so we temporarily ignore the body posture optimization process, which will not affect the propose of our method. \newline\par

\noindent(2) Fault tolerant gait planning\par
\begin{figure}[h] 
\centering 
\includegraphics[scale=0.85]{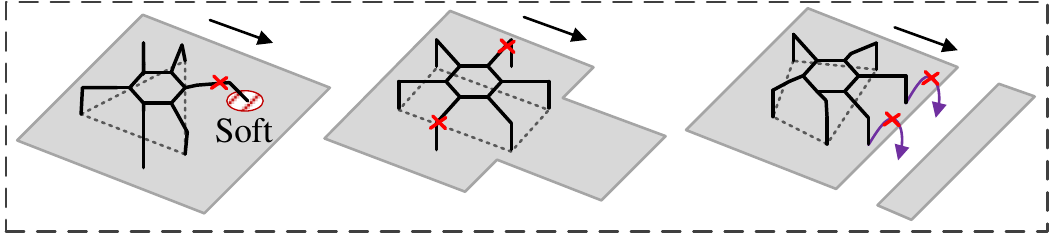} 
\caption{Schematic diagram of environmental fault-tolerant terrain.} 
\label{evironmentFaultFig} 
\end{figure}

Fault-tolerant gait refers to the gait that occurs when each leg of the robot cannot work correctly due to hardware reasons, such as driver failure, motor failure, legs locked. Because the hexapod robot has more number of legs, the robot can still work in the condition of static stability even if one or two legs cannot run. As shown in Figure \ref{evironmentFaultFig},  In the wild environment, there are situations where it is impossible to guarantee that all legs can fall, such as local slippage, subsidence, or sudden narrowing of the terrain. Therefore, referring to the idea of fault tolerance, the above situation is also regarded as a temporary leg fault. In a word, we define the leg without candidate foothold or having physical damages as the fault leg. Combined the idea of dynamic fault tolerance, the robot has a stronger ability to pass and adapt to the environment. When the faults occur, the hexapod can continue to walk by raising the fault legs up.  When the fault is eliminated, the robot restores the function of the faulty leg.

\noindent(3) Planning Method\par
The gait planning in the article relies on heuristic rules because these rules are the only ones that can plan leg motions accurately and guarantee stability using physical laws. The task of free gait is to choose which leg is the swing leg and which leg is the support leg aperiodic during the walk. 

\subsubsection{Support State Planning}
In most cases, there are three criteria to choose support state.\par
One criterion to be taken into account in support state planning is the maximization of the stability margin. It is safer for the robot to have a larger static stability margin during walking.\par
Second,  maximize the amount of robot advance in a step,  which is determined by the step length of the robot.  It can be seen from the formula \ref{MSL},  the maximum step length is determined by the robot's support state,  kinematic margin,  and robot's pose.\par
Third, to realize the idea of fault tolerance, choosing other legs as possible supporting legs instead of selecting the fault leg as the supporting leg. Adding the support states that can make the robot stable in support state table \ref{SupportStateList} to the Set $ S$.\par
To any $s\in S$, the robot has the max step length $MS_{s}$ at $ MD_s $ direction on the condition that the robot has a stable state. For fault-tolerant gait, $MD_s$  represents the direction vector that advances in the support state $s$ and changes continuously during the path tracking process. To any $s\in S$, noting that the robot's stability margin as $SM_{s}$.\par
Based on the above criteria,  the evaluation equation for selecting the support state $s_0$ is designed below:

\begin{equation}
 \left\{\begin{aligned}
         f(s) &=  \omega_1 \cdot MS_{s} + \omega_2 \cdot SM_{s}\\
       s_0 &=\mathop{\arg\max}_{s} (f(s))
       \end{aligned}
\ \right.
 \qquad \text{$s\in S$}
\end{equation}

$\omega_1 $ rewards the maximum step length of the robot at the current state. The larger $\omega_1 $ is, the more likely the robot is to take larger steps and move faster. The $\omega_2 $ rewards the static stability margin of the hexapod robot, the value of $\omega_2 $  increases,  and the robot tends to choose a support state with a larger static stability margin. \par 

According to the support state table \ref{SupportStateList}, the relevant evaluation values of all support states can be calculated, but the candidate states need to be filtered before calculating the evaluation values.\par
$\bullet$ Delete support states with static stability margin less than 0 before state transition. \par
$\bullet$ Delete the state where the fault leg is selected as the supporting leg at the current stage. \par
$\bullet$ Delete the same support state as in the previous step. If it is retained, the program may end up in an infinite loop\newline\par

\subsubsection{Step Length Planning}
The planning of step length also needs to consider the trade-off between the robot’s speed and stability. As long as the stability can be guaranteed,  it is better to plan a longer step length. Because the support polygons are reduced using the method mentioned before, a certain margin of stability margin is reserved to ensure the static stability of the robot. Here we set the step size to the maximum step size $MS_{s_0}$ to maximize the walking speed of the robot. \newline\par

\subsubsection{Foothold Planning}
For each swing leg, there may be multiple alternative footholds in its future workspace pretending that the hexapod's body has moved supported by the combination of supporting legs.
There are two principles for choosing footholds in this article. First, for a specific leg of the robot, it is more inclined to select foothold with higher leg's reduced kinematic margin $KM$. Second, the foothold selection prefers to choose the foothold combinations which make the robot have a higher stability margin. Noting the all possible foothold combinations of swing legs as set $C$, and noting the selected foothold combination of swing legs as $c_0$. The foothold combination selecting method is shown in the following equation.
\begin{equation}
 \left\{\begin{aligned}
         f(c) &=  \omega_L \cdot \overline{KM}(c) + \omega_M \cdot SM(c)\\
       c_0 &=\mathop{\arg\max}_{c} (f(c))
       \end{aligned}
\ \right.
 \qquad \text{$c\in C$}
\end{equation}
$\overline{KM}(c)$ represents that average of all swing legs' kinematic margin after all swing legs swing done according to the foothold combination $c$. $SM(c)$ represents that the static stability margin of the hexapod robot with $c$ foothold combination of swing legs when the hexapod's body move $MS_{s_0}$ at $MD_{s_0}$ direction then reach the next state. The first item rewards the reduced kinematic margin of the foothold. The second item rewards the hexapod robot's static stability margin at the end of the state transition.
$w_L$ and $w_M$ are corresponding weight coefficients, and their values are greater than 0.

For fault legs, the choice of foothold is different. It does not have a fixed foothold. The virtual foothold of the leg moves with the movement of the body and floats in the air.\newline\par

\subsubsection{Trajectory planning}
Given the start and target COG posture and target foothold position, the trajectory of the COG and swing leg needs to be planned to find a smooth trajectory. In addition, the trajectory should also ensure collision-free, optimal energy consumption, etc. In this paper, we use a simple polynomial method to plan the trajectory of the body. By setting constraints such as the position, velocity, and acceleration of the starting point, a trajectory equation with continuous acceleration can be obtained. \newline\par

\noindent(4) Defects of Expert Method\par
$\bullet$ The planning of gait does not consider environmental information, which will affect the plan of footholds.\par

$\bullet$ The planning for the step length of the robot is too violent, which affects the selection of the foothold. The two are coupled with each other.\par

$\bullet$ The selection of the foothold considers the environment where the robot is, but it ignores the future situation. The planning will not only affect the next step planning but also affect all the decisions behind it.\par

The above limitations can be summarized as the planning of gait, step length, and foothold of a legged robot is a sequential decision problem. All the previous decision-making plans will have an impact on subsequent decisions. A well-designed rule-based expert planning method cannot meet all requirements, and there are always some situations that cannot be dealt with.\newline\par

\section{Gait and Foothold Planning Based MCTS}
\subsection{Standard MCTS Method}

\noindent(1)Basic MCTS Algorithm

\begin{figure}[ht] 
\centering 
\includegraphics[scale=0.5]{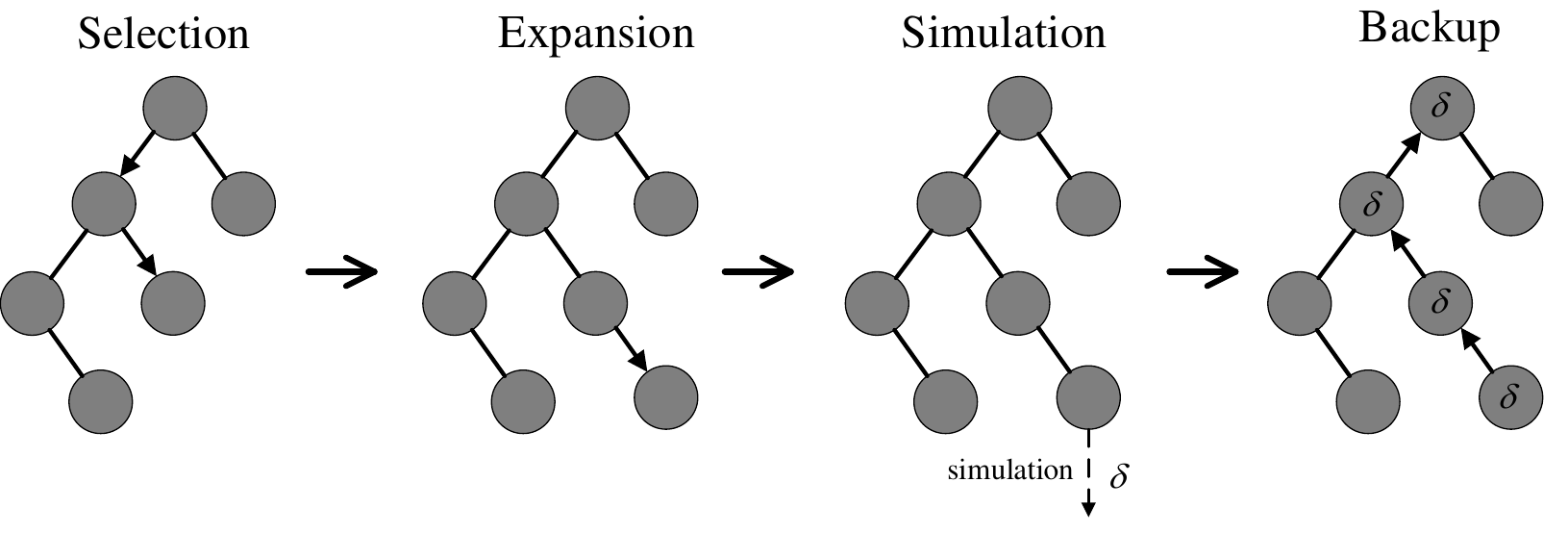} 
\caption{Workflow of  Monte Carlo tree search algorithm} 
\label{stantardMCTS} 
\end{figure}

Monte Carlo Tree Search (MCTS) is an algorithm that uses random (Monte Carlo) samples to find the best decision. Here, we briefly outline the main ideas of MCTS, as is shown in Figure \ref{stantardMCTS}: First, the selection process is based on an existing search tree. Traverse the tree according to the tree strategy to decide which direction to take at each branch until it reaches the leaf node. Then, use one of the remaining possible actions to expand the leaf node to obtain a new leaf node. Starting at that node, within a fixed range or until the final node is reached, a simulation (often also called scaling) will be performed using some default strategy (i.e. the default behaviour of all relevant participants). Update the values of all traversed nodes based on some cost functions that evaluate the simulation results.\par

MCTS algorithm can be grouped into two distinct policies:\par

$\bullet$ Tree Policy: Select or create a leaf node from the nodes already contained within the search tree. Search tree strategies need to strike a balance between exploitation and exploration, the classic method of search tree strategy is UCB (Upper Confidence Bound) algorithm \cite{kocsis2006bandit}.\par

$\bullet$ Simulation Policy: Play out the domain from a given non-terminal node to produce a value estimate.

MCTS has many advantages, which makes it useful for the legged robot to plan its gait, foothold sequence:\par
$\bullet$ MCTS is a statistical anytime algorithm for which more computing power generally leads to better performance. It can be stopped, and a result is available. It might not be optimal but valid.\par
$\bullet$ MCTS can be used with little or no domain knowledge.\par

$\bullet$ MCTS can enforce different policies on different nodes, so it is easy to scale.\par

$\bullet$ MCTS can be highly parallelized, with multiple iterations at a time and multiple simulations at a time.  It facilitates engineering applications.\newline\par

\noindent(2) Extensions for Legged Robot Planning Domain\par
Based on MCTS, this section proposes two modified MCTS methods for hexapod robots, one of which is called the Fast-MCTS method and the other is called as Sliding-MCTS method. First, introduce some definitions for standard MCTS in the field of hexapod robot planning.\par

\begin{figure}[ht] 
\centering 
\includegraphics[scale=0.7]{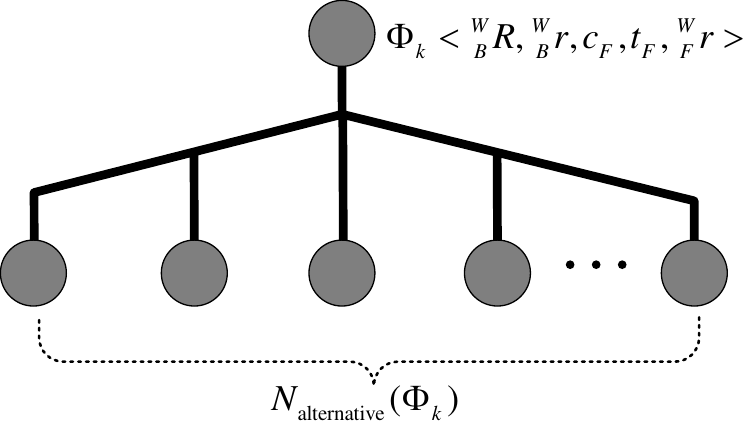} 
\caption{Schematic diagram of alternative nodes} 
\label{alternativeNodeFig} 
\end{figure}

1)Action Space: For hexapod robot planning, each node of the Monte Carlo tree represents the state of the robot, $\Phi:= <_B^W\!R,\  _{W}^{B}\!{r}\ , c_F,\ ,t_F,\  _{F}^{W}\textrm{r}> $, which includes the robot's posture, position, foothold position, support status during the transfer process, leg error status. The set of actions that lead the robot from the current node to the candidate nodes is called the action space. According to Table \ref{SupportStateList} , $n$ alternative support states for a robot state can be obtained and note these $n$ alternative support states as set $S$. For any support state $s\in S$, the maximum advancement $MS_s$ corresponding to the advancing direction $MD_s$ can be obtained. Discrete the maximum advancement $MS_s$  into three different step sizes: $MS_s/3$, $2\cdot MS_s/3$, and $MS_s$, which  constitute the set $L$. For a step length of $l \in L$ and the supporting state s, there are $m_{l,s}$ combinations of footholds. Define the number of candidate states for each hexapod robot state  $\Phi_k$ as $N_{alternative}(\Phi_k)$, as is shown in Figure \ref{alternativeNodeFig} , its calculation formula is as follows:
\begin{equation} \label{alternativeStateNum}
    N_{alternative}(\Phi_k) = \sum_{s \in S}\sum_{l \in L}3m_{s,l}
\end{equation}

2)Search Tree Policy: For search tree policy, we use the standard UCB1 algorithm, the calculation formula is as follows:
\begin{equation}
    UCB1 = X_j + C\cdot \sqrt{\frac{2\cdot{\rm ln}n}{n_j}}
\end{equation}

Where $X_j$ represents the average reward value of node $j$, $ n_j$ is the number of visits to node $j$. $n$ represents the number of visits by the root node. The parameter $C$ is a balance factor, which decides whether to focus on exploration or utilization when select. If the value of $C$ is large, it is more inclined to select some nodes with lower reward value. If the value of $C$ is small, it is more inclined to visit the nodes with higher evaluation reward value. From this, we can see how the UCB algorithm finds a balance between exploration and utilization: both the nodes that have obtained the largest average reward in the past time and the nodes with lower average rewards have chance to be selected.\newline\par

3)Simulation Policy: There are two simulation policies in this paper. The first one is the free fault-tolerant gait planning method introduced in the previous section. The second is an entirely random method, which is to randomly select an executive action in the action space.\newline\par

4)Simulation Horizon: The goal of a game of Gomoku, Go, etc. using Monte Carlo Tree Search is to win. For a hexapod robot walking in a sparse environment, the purpose is to pass it safely. The result returned by the MCTS simulation process of the six-legged robot is set as "Pass" or "Not Pass". In the sparse foothold environment, there are few nodes passed by the simulation, which will lead to node scores mostly 0. When the UCB algorithm is used for selection, the function of utilization is lost. According to the literature \cite{clary2018monte}, humans only planned three steps in advance during the walking process. And it is tested in \cite{matthis2014visual} that planning a certain distance in advance can already get a high passability, and continuing to increase the planning distance has no obvious effect on improving the passability. Therefore, this article sets a simulation view $SH$. If the simulation distance exceeds $SH$, then the simulation result is "Pass".\newline\par

5)Simulation Termination condition: This termination condition applies to both node expansion and simulation. The termination conditions are as follows:\par
$\bullet$ During the continuous $N_{\rm stop}$ state transitions, the robot's forward amount is close to 0.\par
Note: Parameter $N_{\rm stop}$ is set differently for different simulation policies. For example, the value of $N_{\rm stop}$ can be small in the expert method. Because when the robot is stuck,  the possibility of pass using expert method is low. For the random method, the temporary stuck can be released by continually switching the combination of the foothold and the support state, so the value of $N_{\rm stop}$ can be slightly larger.\par

$\bullet$ The expanded node's position is greater than simulation Horizon $SH$ or reaches the target position.\newline\par

6) Node Score and Backpropagation\par
 For each of the node called $j$, the score of it is defined as:
 \begin{equation}
     X_j = N_{{\rm pass},j}  /  N_{{\rm visit}, j}
 \end{equation}

Where $X_j$ represents the score of node $j$, $N_{{\rm pass},j}$  represents the total number of simulation passes for the node  $j$ or the descendants of the node $j$. And $N_{{\rm visit},j}$ represents the total number of visits to node $j$ or its descendants.\par

Using backtracking algorithm to update $N_{\rm visit}(\Phi)$  and $N_{\rm pass}(\Phi)$ from leaf node to root node. Representing $\Phi$ as any of the leaf node's ancestor node. The update formula is as follows:
\begin{equation}
    N_{\rm visit}(\Phi)=N_{\rm visit}(\Phi) + 1
\end{equation}

\begin{equation}
    N_{\rm pass}(\Phi)=N_{\rm pass}(\Phi)+\Delta_{\rm simScore}
\end{equation}
\begin{equation}
\Delta_{\rm simScore}=\left\{
\begin{array}{rcl}
0      &      & {\rm  not \  pass}\\
1     &      & {\rm pass} 
\end{array} \right.
\end{equation}

Where $\Delta_{\rm simScore}$ denotes the result of the simulation of the expanded node. When transfer to the root node, the backpropagation ends.\newline\par


\noindent(3)Defects of Standard MCTS in Hexapod Planning\par
Although the standard MCTS was quickly applied to the field of foot robots, the unmodified standard MCTS has the following problems. \par 
First, the state of each hexapod robot usually has hundreds of candidate states. During the process of building the search tree, the time consumed by the entire expansion increases exponentially. Searching for a state tree passing only 1m has exceeded tens of thousands of nodes, and processing with a single-threaded CPU takes up to ten minutes. This is too unfriendly for real-time planning of foot robots.\par
Second, in a dense foothold environment, it can be passed through a simple expert method, so there is no need to spend a lot of time to use MCTS method.\par
Third, the binary scoring method in measuring node scores is too crude. Although this method can also search for feasible solutions, there is no tendency to optimize the search result sequence. For example, it is more desirable to obtain a faster walking sequence.

\subsection{Selection Planning Based Fast MCTS}
In order to solve the standard MCTS speed problem, this section proposes a fast Monte Carlo search method for the planning of the hexapod robot, and it is called Fast-MCTS. In the simulation step of the standard MCTS method, a large number of simulations have been performed. But only the results of the simulation are used, and the state sequence obtained during the simulation was discarded. The Fast-MCTS uses simulation sequences to quickly build the master branch of the search tree and iteratively updates the master branch by the branch with the highest potential to the destination. The primary purpose of this algorithm is to construct a feasible state sequence quickly, but its optimality cannot be guaranteed. The fast Monte Carlo tree search algorithm is different from the standard MCTS framework. It consists of four main steps:  Extension, Simulation, Updating Master Branch, and Backtracking.\par
First, take the starting state $\Phi _{\rm start}$ of the hexapod robot as the specified starting node$\Phi _{ k}$. \par
$\bullet$ Extension: Expand all candidate states of the specified node $\Phi_k$, each candidate node can only be expanded once. Note the nodes expended as set $AS_{\Phi_k}$\par

$\bullet$ Simulation: To each node ${{\Phi_0}}\in AS_{\Phi_k}$, using the default strategy simulation(Expert method or random method) until reaching the termination condition. Noting the distance of the simulation as $d({\Phi_{0}})$. Taking the nodes of the simulation generated as set $T_{\Phi_0}$. The simulation termination conditions are similar to the standard MCTS simulation, but without the parameter of the simulation horizon. The simulation termination condition of this method is that the robot is continuously stuck or reaches its destination.\par

$\bullet$ Updating Master Branch: Select the extended maximum simulation distance node $\Phi _{k,f} \in AS(\Phi _ k) $.\par
\begin{equation}
    \Phi_{k,f} =\mathop{\arg\max}_{\Phi \in AS_{\Phi _k}} (d(\Phi))
\end{equation}

Adding simulation node sequence $T_{\Phi_0} $ to the search tree and considering the branch as master branch.\par

$\bullet$ Backtracking: If the master branch does not reach the destination, then select the node $\Phi_I$ from the leaf node, which is closest to the target, toward the root node successively, and start to expand, simulate, and update the master branch.\par

\begin{figure}[h] 
\centering 
\includegraphics[scale=0.51]{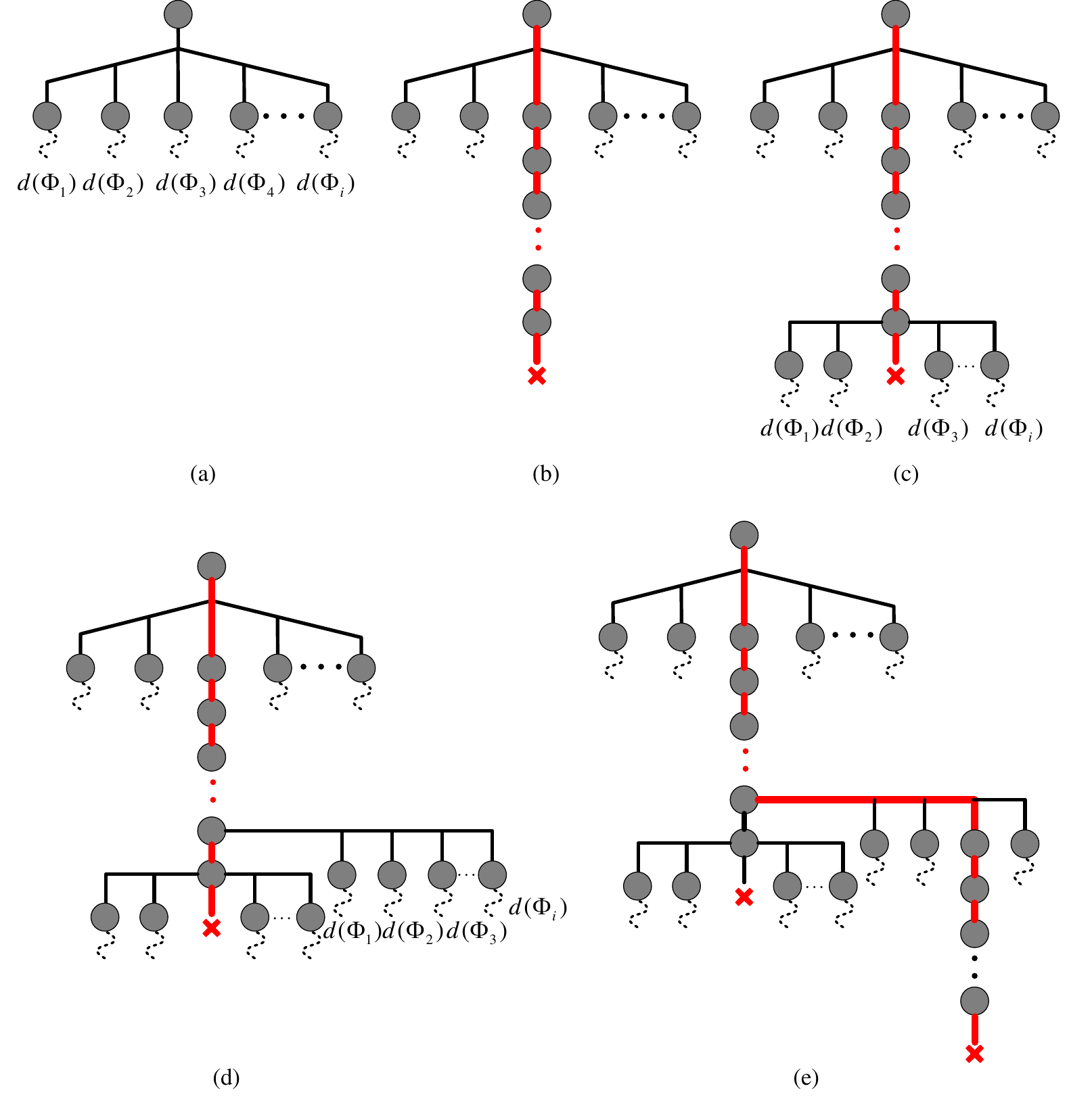} 
\caption{Workflow of Fast-MCTS} 
\label{fastMCTSfig} 
\end{figure}

Next, introduce the flow of the entire algorithm, according to Figure \ref{fastMCTSfig}. In the Figure  \ref{fastMCTSfig}(a), all the candidate state nodes are expanded according to the selected node. Then the simulation is performed with them as the starting point,  the simulation distance and state sequence are recorded. In Figure \ref{fastMCTSfig}(b), the node with the largest simulation distance is selected for expansion, and each node in the state sequence recorded in the simulation is added one by one. The master branch indicated by the thick solid line in the figure. Figure  \ref{fastMCTSfig}(c)(d)(e) indicates that if the master branch does not reach the destination, the algorithm will gradually expand backwards from the furthest child node and update the master branch. The end condition of the entire algorithm is: the tree node reaches the destination, or the program traces back to the root node. The method is presented as pseudocode in Algorithm 1.\par

The algorithm uses the results of the simulation to establish the main branch quickly and updates the main branch by backtracking until it reaches the destination or back to the root node. The idea of the algorithm is to find the position where the robot is easily trapped through multiple simulations, and then keep back and try again until it finds an available solution. The method is consistent with human behaviour during walking. Although this algorithm cannot guarantee the optimality, it has a fast search speed and an excellent effect on improving the passability of a specified strategy and for example, improving the passing performance of expert methods.

\subsection{Selection Planning Based Sliding MCTS}
The two methods of planning a six-legged robot based on MCTS have been introduced. The first method uses the standard MCTS method for planning of a six-legged robot. The entire algorithm is very computationally intensive and time-consuming. Fast-MCTS can quickly find a feasible path, which has a good effect on improving the passability of the expert planning method. However, the entire algorithm is too sparsely sampled, does not highlight the idea of estimating the global situation through sampling, and does not optimize the solution sequence. In view of the above problems, this paper proposes a search algorithm that not only improves the search speed but also optimizes the random sequence. It is defined as Sliding-MCTS.\par
The core processing steps of the algorithm are described below:\par
1): Moving root node\par
Sliding-MCTS is similar to the standard MCTS method. The most crucial difference is that the root node of standard MCTS is fixed, while the root node of Sliding-MCTS will change after a period of sampling.\par
\begin{figure*}[ht] 
\centering 
\includegraphics[scale=0.8]{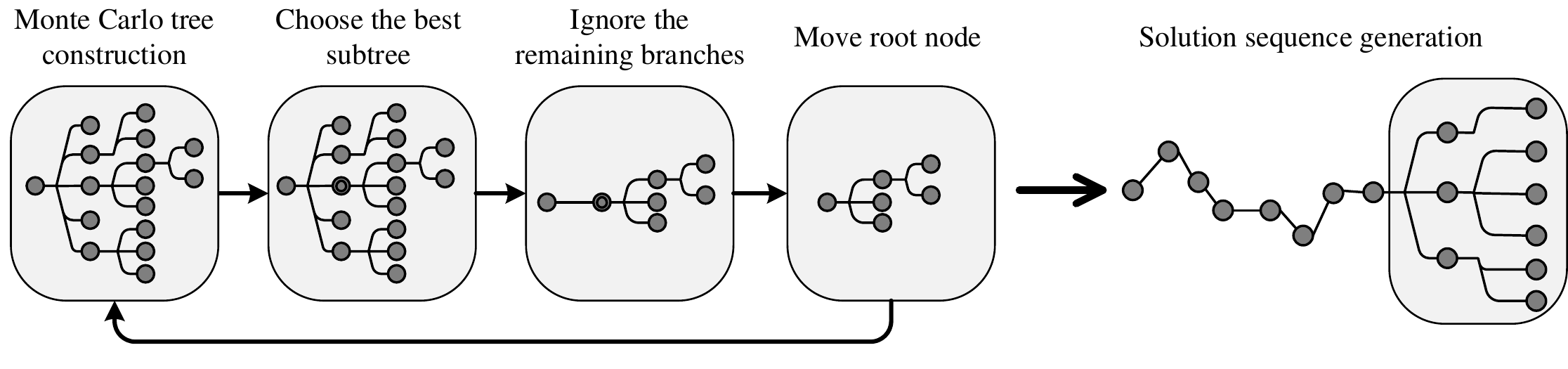} 
\caption{Workflow of sliding-MCTS} 
\label{slidingMCTS} 
\end{figure*}

The core idea of this algorithm is that each step of the robot decision is determined after a large number of samples. Once the best next step in the current situation is selected, then the node corresponding to the state of the robot at this step is chosen as the new root node. As shown in Figure \ref{slidingMCTS}, the simulation in each pane selects the best next step to continue, circularly, a sequence of states is generated.\par

2):Simulation Horizon\par
To facilitate the subsequent quantization of the node score, the simulation horizon distance $SH$ described above is set to a fixed number of simulation steps $N_{\rm SimStepNum}$.\par

3):Node score\par
\begin{figure}[h] 
\centering 
\includegraphics[scale=0.75]{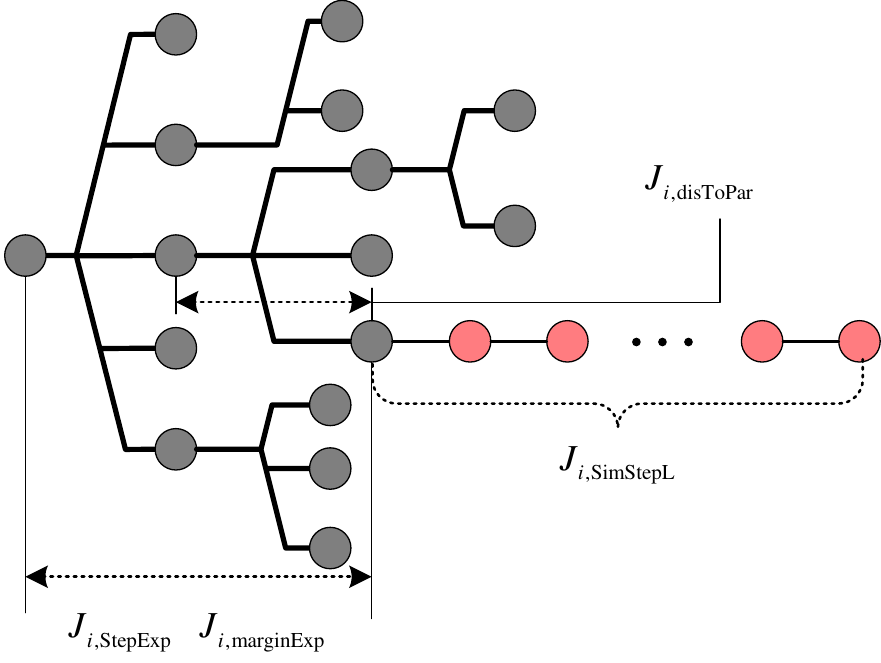} 
\caption{Schematic diagram of the reward function indicator composition. Gray nodes indicate expanded nodes and red nodes indicate simulation nodes.} 
\label{estimateFunctionFig} 
\end{figure}

In the previous article, we introduced the node score, which is defined as the number of successful simulations divided by the number of visits. According to the definition of this score, although an available solution sequence can be found in most cases, there are still some problems. First of all, the definition of scores lacks relevant indicators in the field of legged-robots, which results in the algorithm having no effective target at runtime. Although the solution sequence can be found, people still prefer the algorithm to plan a sequence that walks faster or more stable. In addition, in some cases, some nodes can pass during simulation, but the distance of the node from its parent node is minimal. The algorithm sometimes selects this type of node repeatedly, resulting in an infinite loop and unable to obtain an effective solution. To obtain a higher quality solution sequence, the reward function is used as a new node evaluation method here. The score of node $i$ is defined as $J_i$, and its components are shown in Figure \ref{estimateFunctionFig}. The composition of the score items is as follows:\par

$J_{i,{\rm SimStepL}}$: It rewards the average step size of the simulation sequence from node $i$ with a fixed number of simulation steps $N_{\rm SimStepNum}$. The longer the simulation distance, the larger the average step size of the sequence. The higher the value of this parameter is, the higher the potential for the node $i$ to have greater passability\par
$J_{i,{\rm StepExp}}$: It rewards the average step size of the extended sequence nodes from node $i$ to the current root node, making the algorithm tend to converge to sequences that walk faster. Define the state sequence from the extended node $i$ to the root node as a set $C_i$. Define the quantity of elements in $C_i$ as $n_i$. Take the step size between node $i$ and its parent as $s_i$. For the root node $r$, $s_r$ is equal to 0. The formula for calculating $J_{i,{\rm StepExp}}$ is as follows:\par
\begin{equation}
J_{i,{\rm StepExp}} = \frac{1}{n_i}\sum_{j\in C_i }s_j
\end{equation}
$J_{i,{\rm marginExp}}$: It rewards the average static stability margin of the extended sequence nodes from node $i$ to the current root node, making the algorithm tend to converge to a sequence with a larger average stability margin. The $SM_i$ represents the stability margin of node $i$. The formula for calculating $J_{i,{\rm marginExp}}$ is as follows:\par
\begin{equation}
J_{i,{\rm marginExp}} = \frac{1}{n_i}\sum_{j\in C_i }SM_j
\end{equation}
$J_{i,{\rm disToPar}}$: It rewards the step size from the node $i$ to its parent node, preventing the algorithm from repeatedly accessing nodes with a minimal forward distance.\par
The sum of each term multiplied by the corresponding weight is the score of node $i$.\par
\begin{equation}
    J_i=\sum \omega_{i,(.)}J_{i,(.)}
\end{equation}
Where $\omega_{i,(.)}$ represents the weight coefficient of each term. According to our debugging experience, the weight value corresponding to $J_{i,{\rm SimStepL}}$ and $J_{i,{\rm StepExp}}$ can be larger. $J_{i,{\rm disToPar}}$'s weight value can be small to prevent getting stuck in advance due to excessively greedy forward speed.\par

4):Score Backpropagation\par
When the extended node calculates a new reward score, upward propagation does not calculate the average of all extended node scores but retains the highest score. The propagation formula is as follows:
\begin{equation}
    X_i=J_i
\end{equation}

\begin{equation}
X_j=\left\{
\begin{array}{rcl}
J_i       &      & {\mathrm{if}  J_i>X_j}\\
X_j     &      & {\mathrm{else}} 
\end{array} \right.
\end{equation}

For the gait and foothold sequence planning of a legged robot, the goal is to find only one result sequence. Therefore, it is better to measure the quality of a tree by its best child nodes. Conversely, if the average score of the entire tree is used as a measurement index, some nodes with lower scores will diminish the scores of the best nodes. In a sparse foothold environment, there are a few solution sequences that can pass, and such a measure will make it difficult for the algorithm to find these solutions.\par

5):Single Step Decision Time\par
The state of the robot at each step is determined after a certain period of sampling. Define the single-step decision time as the time required for $N_{\rm Samp}$ samplings. The parameter $N_{\rm Samp}$ can be adjusted according to the actual situation. As the complexity of the environment increases, the value of $N_{\rm Samp}$  can increase correspondingly.

6):Algorithm Termination Condition\par

\begin{figure}[htb] 
\centering 
\includegraphics[scale=0.5]{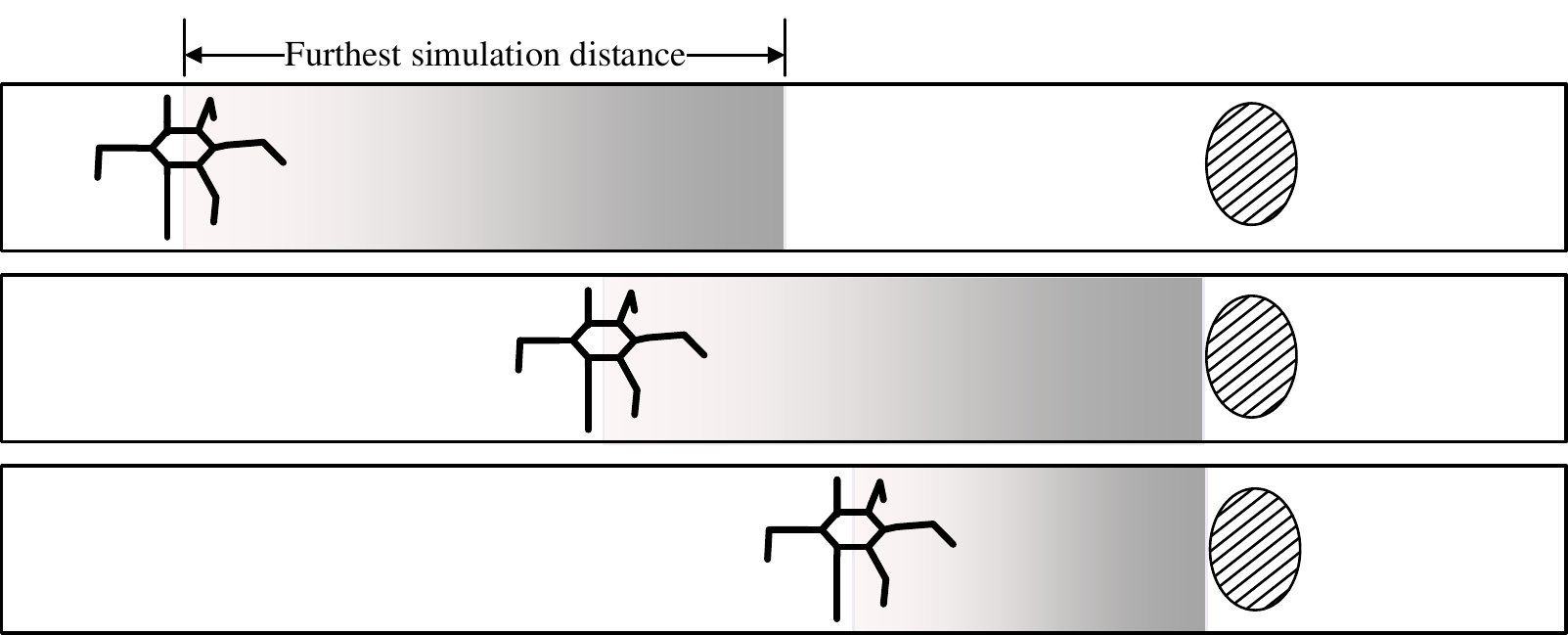} 
\caption{Schematic diagram of the movement of the simulation horizon.} 
\label{simulatioinHorizon} 
\end{figure}

There are two algorithm termination conditions: first, if the extended node reaches the specified target point, the algorithm stops; second, if the expanded node approaches the farthest simulation distance, the algorithm terminates. It happens when encountering an area that cannot pass. As shown in Figure \ref{simulatioinHorizon}, the edge of the grey area is the farthest simulation position of the robot. If the farthest simulation position is very close to the current expansion node, the entire algorithm cannot continue.\par

7):Choose the best subtree\par
The best subtree is selected using the standard UCB formula, but the coefficient $C$ is set to zero. Select the branch where the node with the highest score is located, and subtract the remaining branches.

Although Sliding-MCTS does not completely optimize the entire sequence, the algorithm still has a good effect. There are three reasons for this. First,  as mentioned earlier,  only planning certain steps in advance will hardly reduce the overall passability. Second, compared with the standard MCTS algorithm, Sliding-MCTS can greatly decrease the search time. Third,  based on the parameter $N_{\rm Samp}$ ,$N_{\rm SimStepNum}$and simulation steps $N_{\rm SimStepNum}$, the search time and the optimization degree can be effectively balanced. The method is presented as pseudocode in Algorithm 2.\par

\section{Experiment}
\begin{figure}[h] 
\centering 
\includegraphics[scale=1]{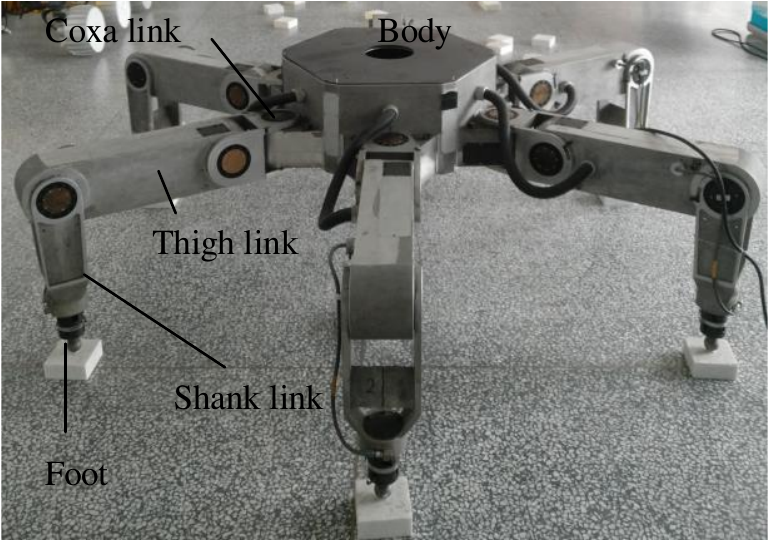} 
\caption{Elspider hexapod robot} 
\label{ELSpiderFig} 
\end{figure}

We validated our approach on the Elspider robot\cite{liu2018static}\cite{liu2018state}\cite{gao2019low}. The experimental platform is an electric heavy-duty hexapod Elspider developed by Harbin Institute of Technology, as shown in Figure \ref{ELSpiderFig}. The overall mass of the robot is 300kg, and it can walk stably under a load of 150kg. The design of the machine adopts a high-stability uniformly distributed six-leg configuration, and the driving wheelsets are evenly distributed at the base joints of each leg.  The robot is approximately 1.9m long, 2.1m wide, and 0.5m tall. The relevant parameters of the robot are shown in Table \ref{robotParameter}, the radius of the trunk body(0.4m), coxa link(0.18m), thigh link(0.5m), and shank link(0.5m).\par

\begin{table}[]
\centering
\caption{Mechanical and geometric parameters of Elsiper robot}
\label{robotParameter}
\begin{tabular}{ccc}
\hline
Parameter  & Lengh(m) & Mass(kg) \\
\hline
Body       & 0.4      & 121.9    \\
Coxa link  & 0.18     & 3.6      \\
Thigh link & 0.5      & 22       \\
Shank link & 0.5      & 7.2      \\
Foot       & 0.025    & 0.2        \\
\hline
\end{tabular}
\end{table}

\begin{table*}[ht]
\centering
\caption{Experimental parameter setting table}
\label{parameterForAlgorithm}
\begin{tabular}{lll}
\hline
\multicolumn{1}{c}{\textbf{Parameter}}            & \multicolumn{1}{c}{\textbf{Value}} & \multicolumn{1}{c}{\textbf{Parameter Meaning}}  \\
\hline
$BM_{\rm min}$  & 0.05{\rm m}     & Minimal static stability margin                \\
$\omega_1$               & 0.7     & Support state planning weight coefficient                \\
$\omega_2$               & 0.3     & Support state planning weight coefficient                \\
$\omega_L$               & 0.7     & Foothold planning weight coefficient                \\
$\omega_M$               & 0.3     & Foothold planning weight coefficient                \\
$N_{\rm stop}$               &5     & Threshold for the number of consecutive stuck                \\
$N_{\rm SimStepNum}$                &20      & Fixed simulation steps                \\
$\omega_{i,{\rm SimStepL}}$               &3     & Sequence evaluation function weight coefficient                 \\
$\omega_{i,{\rm StepExp}}$               &1      & Sequence evaluation function weight coefficient                \\
$\omega_{i,{\rm marginExp}}$               &0.5      & Sequence evaluation function weight coefficient                \\
$\omega_{i,{\rm disToPar}} $              &0.2     & Sequence evaluation function weight coefficient                \\
$N_{\rm Samp}$               &500     & Number of single steps sampling for sliding MCTS                \\
$C$               &0.3     & UCB algorithm balance coefficient                \\

\hline
\end{tabular}
\end{table*}

To examine the behaviour of the proposed algorithm, We designed three different types of experiments to expand the description. The first experiment is performed on terrain with randomly distributed footholds. This experiment can statistically compare the different planning methods' ability to pass complex terrain, speed of advance, and planning time. By reducing the support polygon area, the stability of the robot is ensured. Therefore, the comparison of the body stability margin index is not performed in the experiment. The second type of experiment is tested in artiﬁcially designed challenging terrains to verify the validity of the proposed method. The last experiment is to test on a real robot to illustrate how the proposed method can be applied to a real environment.\par

The experimental planning method includes the following six methods: (1) Triple gait.  (2) Wave gait. (3) Free fault-tolerant gait. (4) Fast-MCTS adopts a random scheme as a simulation strategy, which is defined as Fast-MCTS (Random). (5) Fast-MCTS adopts the free-fault-tolerant gait expert scheme as the planning scheme of the simulation strategy, which is defined as FastMCTS (Expert). (6) Sliding-MCTS method,  its simulation strategy uses a random method.\par

Triple gait and wave gait are two typical gait methods commonly used by hexapod robots. The diagonal gait is the fastest, while the wave gait is the slowest but the most stable. The planning of step length and foothold is the same as the expert planning method described above. If the robot is trapped or reaches the target point, the algorithm ends. \par 

The latter three methods are planning methods based on MCTS. As mentioned in formula \ref{alternativeStateNum}, there are $\sum_{s \in S}\sum_{l \in L}3m_{s,l}$ candidate states of state $k$. To reduce the calculation amount of the algorithm, only one foothold combination is reserved through expert planning method for a support state. Therefore, the number of candidate states for the next step of each state is reduced to $\sum_{s \in S}3$.  How to select valuable alternative states to accelerate search time is also a research direction.\par

All algorithms run on the Intel i5 2.20GHz notebook computer and only use single-thread programming. The setting values of the entire algorithm parameters are shown in Table \ref{parameterForAlgorithm}.


\subsection{Random Terrain Simulation Experiment}
\begin{figure}[h] 
\centering 
\includegraphics[scale=0.5]{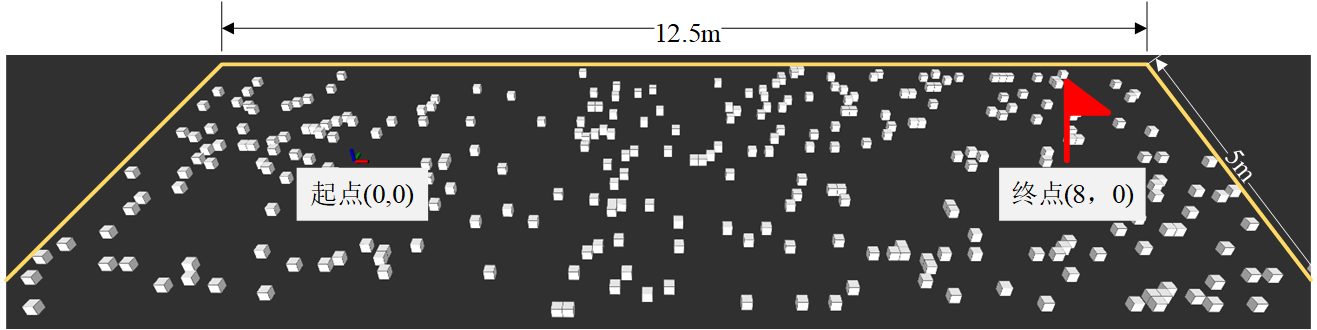} 
\caption{Random foothold distribution experiment map} 
\label{experimentTerrainFig} 
\end{figure}

The terrain of the random experiment is shown in Figure \ref{experimentTerrainFig}. The entire map is 12.5 meters long and 5 meters wide, and the footholds are randomly distributed in this area. The starting point of the robot is the coordinate origin, the forward direction is the positive direction of the $x$ axis, and the target point is (8,0). When the robot advances more than 8 meters, the robot reaches the target point.\par

Experiments were carried out on terrains with three different numbers of footholds, including 400 footholds, 350 footholds, and 300 footholds. Each density generates 20 different maps, and experiments on six different planning schemes are performed on each map.\par

\begin{figure}[H] 
\centering 
\includegraphics[scale=0.6]{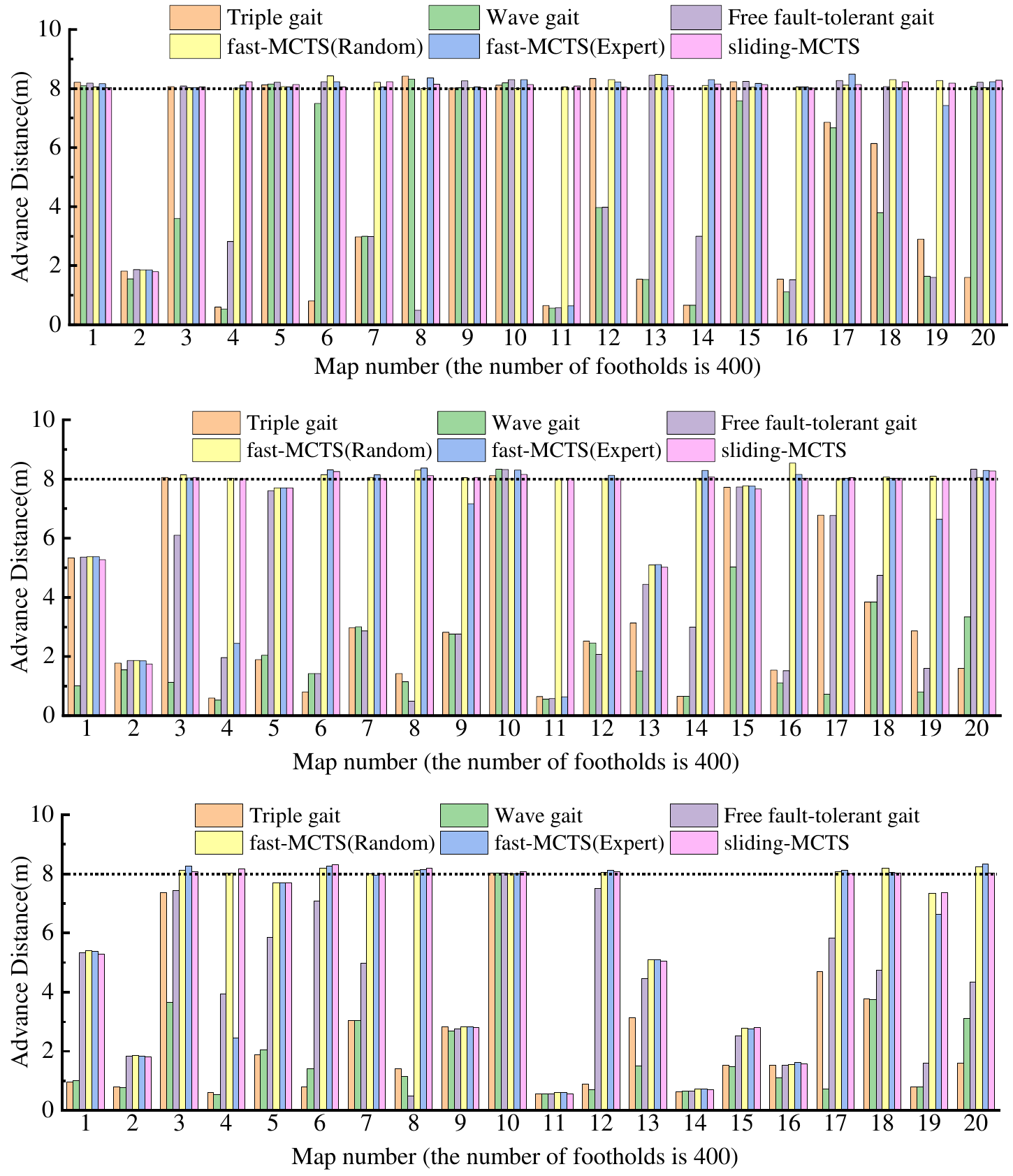} 
\caption{Comparison of advance distance for different planning method tested on different maps.} 
\label{originDataDisFig} 
\end{figure}

Figure \ref{originDataDisFig} shows the raw data of 60 experiments. The abscissa is the label of the different test map, and the ordinate is the distance travelled by the robot. It can be seen intuitively that the passing capacity of the three planning methods based on sequence optimization is much higher than the results of three single-step optimization expert planning method. With the reduction of the number of footholds, the situation that the robot can reach the target point gradually decreases. For the single-step optimization method, it can be seen that in most cases, the free fault-tolerant gait has a larger amount of progress, but there are still some cases where the triple gait goes further.  Although the free fault-tolerant gait method constructed according to expert experience improves the passing ability to a certain extent, the method still cannot guarantee that all cases are better than other typical gait methods.\par

\begin{figure*}[t] 
\centering 
\includegraphics[scale=1]{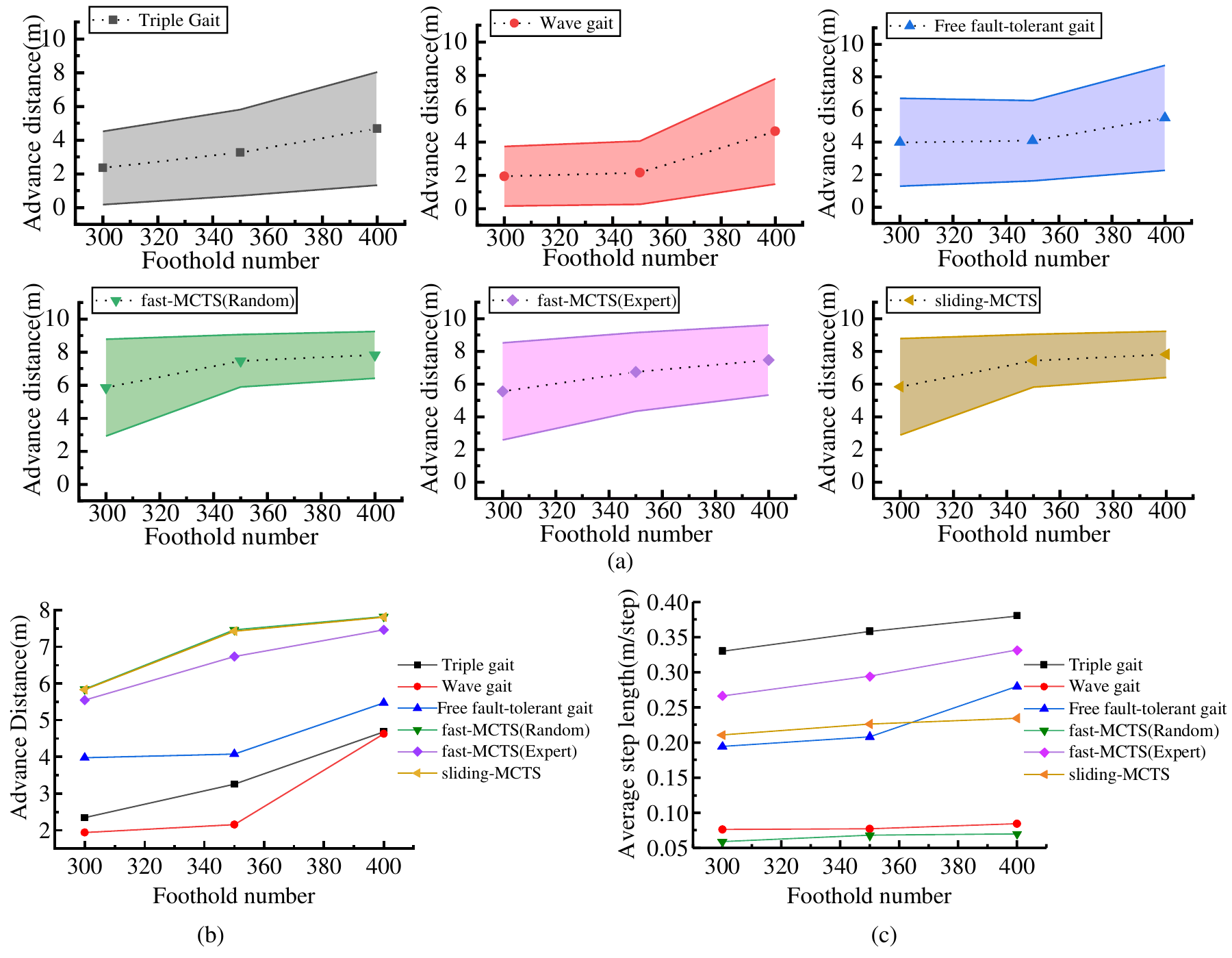} 
\caption{Experimental data of different planning methods under different foothold density environments:(a)Forward distance error band diagram for different planning methods. (b)The comparison chart of the average advance distance represents the passing ability of the robot. (c)Comparison of the average step size of the robot. This can indicate the robot's advancing speed. } 
\label{aaaaaaaaaa} 
\end{figure*}

By statistical analysis of the passability data, the error band diagrams of different planning methods under three different foothold density can be obtained. It can be seen from Figure \ref{aaaaaaaaaa}(a) that as the density of the footholds of the terrain increases, the advance distance of all planning methods gradually increases. The error bands of triple gait, wave gait and free fault-tolerant gait become broader as the density of the foothold increases. In contrast, the error bands of the last three planning methods using MCTS gradually becomes narrower as the density of the foothold increases. In the environment with low foothold density, the rule-based expert method has poor passability. Therefore, in most cases, the robot has a small travel distance. The increase in the density of footholds has improved the robot's passability. However, there are still some maps that the robot cannot pass due to the defeats of the rule-based method. Therefore, the result shows that the error band becomes wider with the rise of the foothold density of terrain. MCTS-based sequence optimization methods are different. Most of the latter three planning methods can still travel long distances in low foothold density environment, and a small part cannot pass because the environment is too harsh. When the foothold density of terrain increases, the robot can reach the destinations almost for all maps. Therefore, the result shows that the error band becomes narrower with the rise of the foothold density of terrain.\par

Compare the passing capabilities of all planning methods, as shown in Figure \ref{aaaaaaaaaa}(b).  It can be seen that the free fault-tolerant gait has significantly higher passing capacity than the diagonal gait and wave gait. In terms of passing ability, the three planning methods using sequence optimization are far superior to the first three methods. The passing capacity of sliding-MCTS and fast-MCTS is the best.

In terms of forward speed, we use the average length of the entire planning sequence to represent the forward speed of the robot. According to Figure \ref{aaaaaaaaaa}(c), the diagonal gait is the fastest, and Fast-MCTS (Expert) is the second-fastest one. The free fault-tolerant gait has a slower walking speed than Sliding-MCTS in a sparse foothold environment, and the walking speed is faster than free fault-tolerant gait when the foothold density of terrain is denser. It can be seen that the Sliding-MCTS method can ensure the best passing ability, besides it can search for a high-speed gait sequence in a sparse foothold environment. The slowest speeds are diagonal gait and Fast-MCTS (Random). Although Fast-MCTS (Random) has a high passing capacity, the sequence it searches for has not been optimized by a large number of samples, resulting in many invalid states in the entire sequence and the lowest speed.

\begin{figure*}[ht] 
\centering 
\includegraphics[scale=0.6]{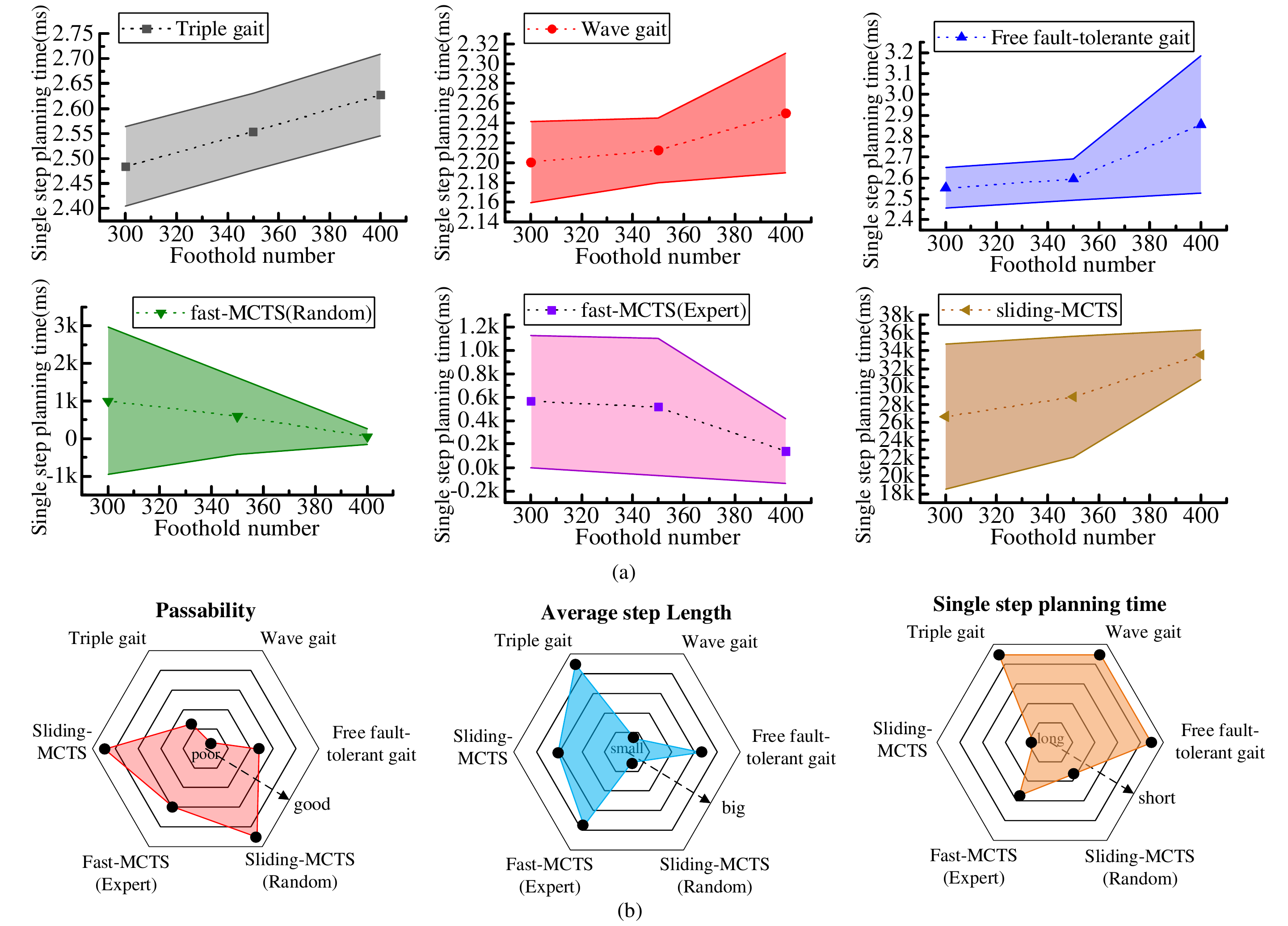} 
\caption{(a)Single-step planning time error band diagram for different planning methods. (b)Index comparison chart of different planning methods} 
\label{bbbbbb} 
\end{figure*}

To compare the planning time of the algorithms, we use the single-step planning time of the entire sequence to represent. As shown in Figure \ref{bbbbbb}(a), the gait planning time of the first three expert methods is about 3ms, and the planning time increases as the density of the foothold increases. It is because when the density of foothold becomes larger,  there are more available foothold can be selected, and more support states can be chosen by free fault-tolerant gait. And it takes more time to calculate with the increase of foothold density.  The single-step planning time of Fast-MCTS algorithm is about 1s. As the environment becomes harsher, the search time gradually increases. For certain sparse foothold environments, the algorithm occasionally finds solution sequences quickly. So this leads to a larger error band for planning time in environments with low foothold density.  The search time of the Sliding-MCTS method is determined by the number of expansion nodes $N_{\rm sliding}$  planned for each step and the fixed number of simulation steps $N_{\rm SimStepNum}$. Its single-step planning time is about 30s. The reason that the planning time becomes longer as the foothold density increases is the same as the free fault-tolerant gait. In a low-density foothold environment, the number of invalid nodes is unstable. The higher the density of environmental footholds, the less the number of invalid node expansions. This is also the reason why the error band gradually narrows.\par

In summary, as shown in Figure \ref{bbbbbb}(b), the six planning methods have their advantages and disadvantages. In terms of passability, Fast-MCTS (Random) and Sliding-MCTS methods have the highest passability. Expert planning methods have very poor passability, and free fault-tolerant gaits that take into account environmental fault tolerance have slightly better passability. In terms of walking speed, the triple gait is the fastest, followed by Fast-MCTS (Expert), and the speeds of free fault-tolerant gait and Sliding-MCTS are both fast. The other two methods are slow. In terms of planning time, the three planning methods using MCTS take longer, and the Fast-MCTS single-step planning time is about one second. The single-step planning time of Sliding-MCTS is relatively long, and it is related to its set parameters. However, the planning of each step of this method is independent and is not affected by subsequent plan. Therefore,  Sliding-MCTS is suitable for local planning.

\subsection{Special Terrain Simulation Experiment}

\begin{figure*}[ht] 
\centering 
\includegraphics[scale=0.67]{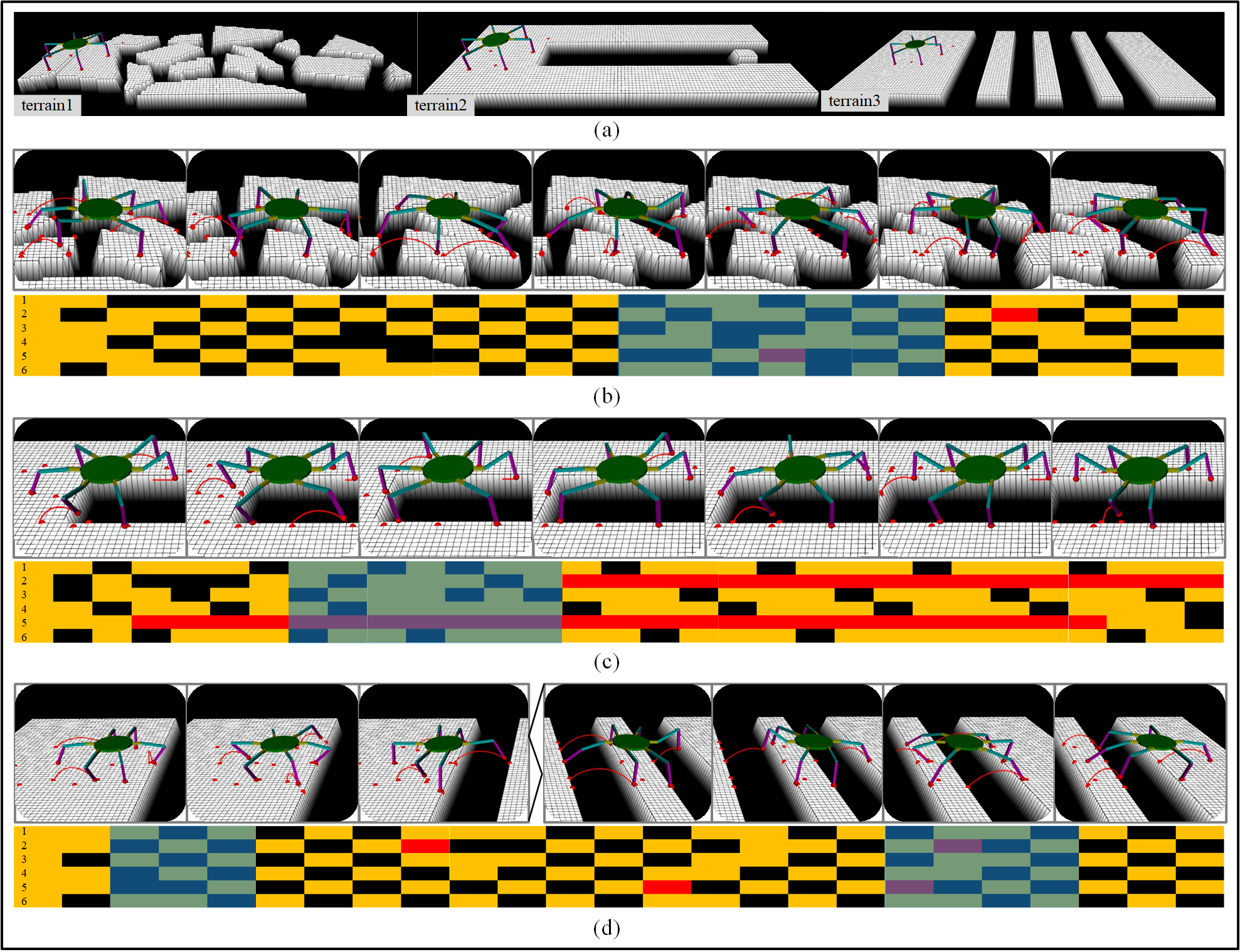} 
\caption{(a) Artificially designed terrains.  (b)(c)(d) Screenshots and gait charts of the robot's passage through three different terrains. The light blue part of the gait diagram corresponds to the screenshot of the robot walking. Each screenshot of the robot represents an operating state, and the red curve is the trajectory of the swinging leg at this time. In the gait diagram, black indicates that the leg is a swinging leg, yellow indicates that the leg is a supporting leg, and red indicates that the leg is a fault-tolerant leg.} 
\label{demoFig} 
\end{figure*}

The proposed sliding-MCTS algorithm is applied to some artificial terrain to verify its validity. As shown in Figure \ref{demoFig}(a), we design 3 different terrains. The first type represents the segmented terrain that can be seen in real life. The second terrain is more extreme, with a rectangular area deducted in the middle of the flat terrain. The third type of terrain represents continuous trench terrain, and the width of the trench is inconsistent. All three types of terrain robots can pass smoothly. Figure \ref{demoFig} is a screenshot of a part of the robot passing the terrain and a gait diagram of the entire process. Figures \ref{demoFig} (b)(c) show that in a harsh environment, the robot can temporarily lift its legs without an effective foothold through such terrain. In Figure \ref{demoFig} (c), the robot even becomes a quadruped robot to cross the terrain. In Figure \ref{demoFig} (d), the robot continuously adjusts the step size to cross the continuous trench terrain effectively. In the gait diagram, black represents the swing state, yellow represents the support state, and red represents the wrong leg (still belongs to swing state). It can be seen that the robot can successfully pass these challenging terrains by continuously adjusting the gait.

\begin{figure*}[ht] 
\centering 
\includegraphics[scale=1]{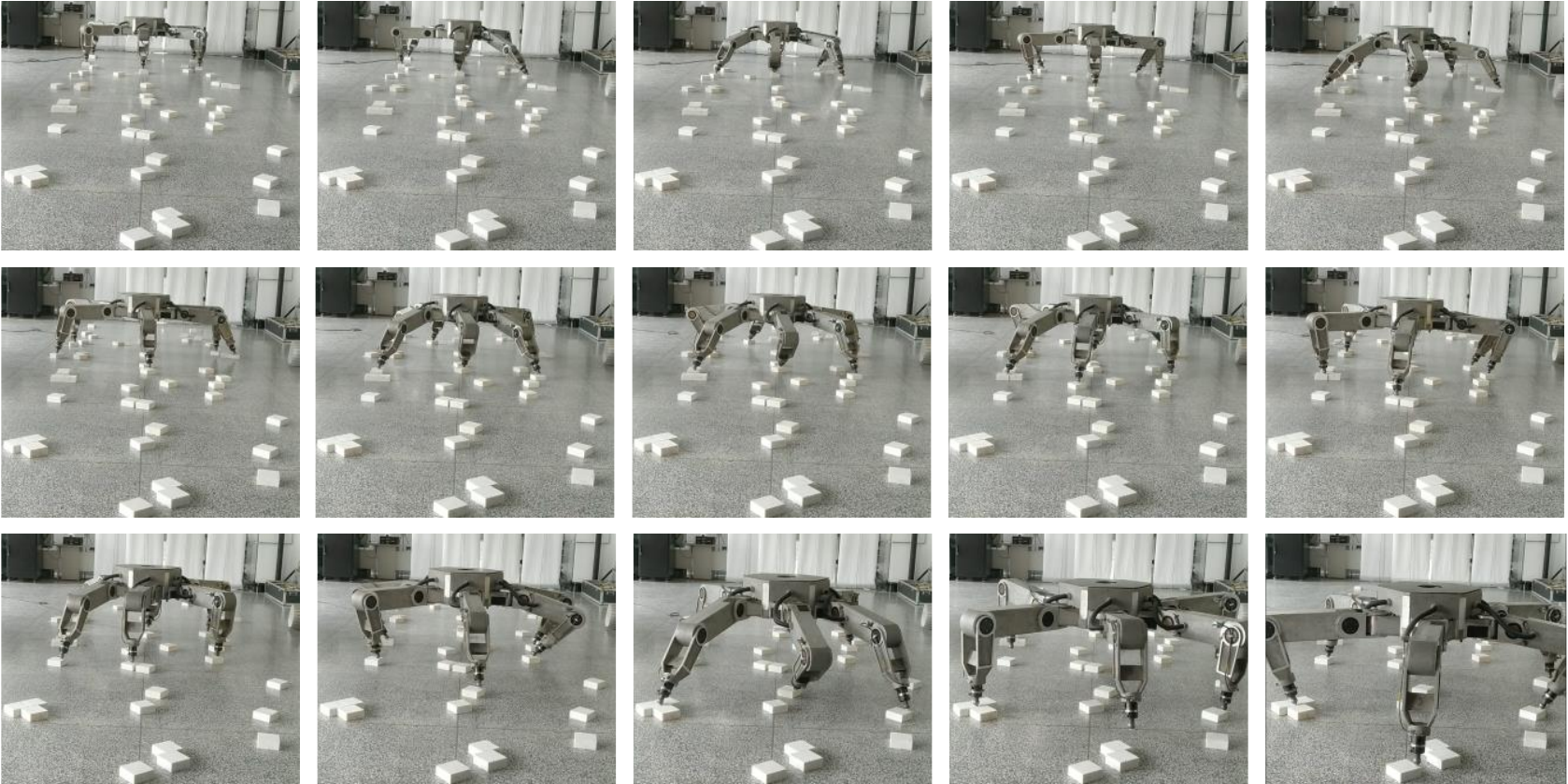} 
\caption{Elspider walking on discrete bricks.} 
\label{realRobotExperiment} 
\end{figure*}

\subsection{Physical robot experiment}
	We carried out some physical experiments to illustrate the feasibility of the algorithm. The experiment terrain is set in advance, as shown in Figure \ref{realRobotExperiment}, bricks represent discrete areas where the robot's feet can land. The position of the robot is measured by a visual capture system. The planning method uses the Sliding-MCTS algorithm proposed in this paper. The robot needs to go straight from one side of the field to the end of the field. It can be seen that the robot can choose the bricks scattered on the ground as a foothold, and successfully reached the target position. The experimental results show that the algorithm proposed in this paper can be effectively applied to physical robots.

\section{Conclusion} 
In this work, a gait foothold planning method based on MCTS is proposed. Before introducing the sequence optimization method, according to the harshness of the environment,  we combine fault-tolerant gait planning and free gait, a free fault-tolerant gait method based on the expert planning method is proposed. According to the particularity of the application of MCTS in the field of legged robots, we have made some changes to the standard MCST and introduced two methods, Fast MCTS and Sliding MCTS. FastMCTS can quickly improve the passing ability of the default planning method, but it is not very convergent. SlidingMCTS can effectively balance the search time and convergence while having a good passing ability. The simulation experiments verify the advantages and disadvantages of different methods,  the rule-based expert method has a fast calculation speed, while the optimization-based method has better passability. The calculation time of the optimization method can also meet real-time requirements. Finally, through artificially designing some challenging terrain to test the algorithm, and applying the algorithm on the physical robot, the results show that the proposed method can have a good passability in the sparse foothold environment. In the future, we will also continue to study how to increase the search speed of the algorithm and combine it with machine vision to explore the wild environment in real-time.

\section{Acknowledgements} 
This study was supported in part by the National Natural Science Foundation of China(91948202), and the National Key Research and Development Project of China(2019YFB1309500).



\bibliographystyle{unsrt}









\newpage

\bibliography{references}

\begin{algorithm*}[!h]
	\caption{Fast Monte Carlo tree search algorithm}
	\begin{algorithmic}[1]
        \Require Initial robot state $\Phi_0$, target position $P_{\rm destination}$
        \Ensure Solution sequence $\Psi=\left \{ \Phi_1,\Phi_2...\Phi_k \right \} $
            \Function {FastMctsSearch($\Phi_0$, $P_{\rm destination}$)}{}
                \State Create the root node $N_{\Phi_0}$
                \State Set the expansion node $N_{\Phi_{\rm expand}} = N_{\Phi_0}$
                \State Set the furthest node of the master branch $N_{\Phi_{\rm end}}$
                \State create variable $disMax = 0$
                \While {"Node $N_{\Phi_{\rm end}}$ has not reached the destination or extended node $N_{\Phi_{\rm expand}}$ is not  the root node."}  
                    \State $S_{\rm sequence}$, $dis$ = \Call{ExpandAndSimulation}{$N_{\Phi_{\rm expand}}$}
                    \If{$dis > disMax$}
                        \State $N_{\Phi_{\rm end}}$ = \Call{UpdataMasterBranch}{$N_{\Phi_{\rm expand}}$,$S_{\rm sequence}$}
                        \State $disMax = dis$
                    \EndIf
                    \State $N_{\Phi_{\rm expand}}$ = \Call{TraceBack}{$N_{\Phi_{\rm end}}$}
                \EndWhile
               \State
                \Return State sequence $\Psi=\left \{ \Phi_1,\Phi_2...\Phi_k \right \} $ between $N_{\Phi_0}$ and $N_{\Phi_{\rm end}}$.
            \EndFunction
            
            \State
            \Function {ExpandAndSimulation($N_{\Phi}$)}{}
                \State Create a sequence storage variable $sequenceList$
                \State Create a distance storage variable $disList$
                \For{each $N_{\Phi_i}$ in the alternative nodes of $\Phi$}  
                  \State Add child node $N_{\Phi_i}$ to $N_{\Phi}$
                  
                  \State $dis$, $S_{\rm sequence}$ = \Call{Simulation}{$N_{\Phi_i}$}
                  \State $disList$.add($dis$)
                  \State $sequenceList$.add($S_{sequence}$)
                \EndFor  
                \State $index$ = $\underset{i}{\rm arg\ max}(disList)$
                \State
                \Return  $sequenceList[index]$, $dis[index]$
            \EndFunction

            \State
            \Function {Simulation($N_{\Phi}$)}{}
                \State $N_{\Phi_{\rm tmp}} = N_{\Phi}$
                \State Create a node sequence storage variable $S_{\rm sequence}$
                \While{Node $N_{\Phi_{\rm end}}$ has not reached the destination or $C_{\rm count} > N_{\rm SimStepNum}$}
                    \State Generate the next state $\Phi_{\rm next}$ of $\Phi_{\rm tmp}$ by expert algorithm or random algorithm.
                    \If{Distance between $\Phi_{\rm tmp}$ and $\Phi_{\rm next} < 0.01$m}
                    \State $C_{\rm count} = C_{\rm count} + 1$ 
                    \EndIf
                    \State $N_{\Phi_{\rm tmp}} = N_{\Phi_{\rm next}}$
                    \State $S_{\rm sequence}$.add($ N_{\Phi_{\rm tmp}} $)
                \EndWhile
                \State
                \Return the forward distance of $N_{\Phi_{\rm tmp}}$, $S_{\rm sequence}$
            \EndFunction

            \State
            \Function {UpdataMasterBranch($N_{\Phi}$, $S_{\rm sequence}$)}{}
                \For{each $N_{\Phi_i}$ in $S_{\rm sequence}$}  
                  \State Add child node $N_{\Phi_i}$ to $N_{\Phi}$
                  \State $N_{\Phi}$ = $N_{\Phi_i}$
                \EndFor  
                \State 
                \Return $N_{\Phi}$
            \EndFunction

    \algstore{myalg}
	\end{algorithmic}
\end{algorithm*}

\begin{algorithm*}[ht]
	\begin{algorithmic}[1]
	    \algrestore{myalg}
            \Function {TraceBack($N_{\Phi}$)}{}
                \While{The number of alternate states of $N_{\Phi}$ is not 0, and $N_{\Phi}$ is not expanded} 
                
                \State $N_{\Phi}$ = the parent node of $N_{\Phi}$
                \EndWhile
                
                \State
                \Return $N_{\Phi}$
            \EndFunction

	\end{algorithmic}
\end{algorithm*}

\begin{algorithm*}[!h]
	\caption{Sliding Monte Carlo tree search algorithm}
	\begin{algorithmic}[1]
        \Require Initial robot state $\Phi_0$, target position $P_{\rm destination}$
        \Ensure Solution sequence $\Psi=\left \{ \Phi_1,\Phi_2...\Phi_k \right \} $
            \Function {SlidingMctsSearch($\Phi_0$, $P_{\rm destination}$)}{}
                \State Create the root node $N_{\Phi_0}$
                \State $N_{\Phi_{\rm expand}} = N_{\Phi_0}$
                \State Create a furthest simulation distance variable $L_{\rm max}$

                \State Create a sliding root node sequence storage variable $S_{\rm sequence}$
                \While {"Node $N_{\Phi_{\rm expand}}$ has not reached the destination or $($the foward distance of $N_{\Phi_{\rm expand}}$ $-$ $L_{\rm max}) < $  threshold"}  
                    \State Create variable $Count = 0$
                    \While{$Count < N_{\rm Samp}$}
                        \State $Count = Count + 1$
                        \State $\Phi_{\rm expanded}$ = \Call{TreePolicy}{$N_{\Phi_0}$}
                        \State $J_{(.)}$, $L_{\rm tmp}$ = \Call{Simulation}{$N_{\Phi_{\rm expand}}$}
                        \If{$L_{\rm max} < L_{\rm tmp}$}
                            \State $L_{\rm max}$ = $L_{\rm tmp}$
                        \EndIf
                        \State  \Call{BackUpdate}{$N_{\Phi_{\rm expand}}$, $J_{(.)}$}
                    \EndWhile
                    \State $N_{\Phi_{\rm best}}$ = \Call{GetBestChild}{$N_{\Phi_{\rm expand}}$, $J_{(.)}$}
                    \State $S_{\rm sequence}$.add($N_{\Phi_0}$)
                    \State $N_{\Phi_0}$ = $N_{\Phi_{\rm best}}$
                \EndWhile
               \State
                \Return State sequence $\Psi=\left \{ \Phi_1,\Phi_2...\Phi_k \right \} $ in $S_{\rm sequence}$
            \EndFunction
            
            \State
            \Function {TreePolicy}{$N_{\Phi}$}
            \While{1}
                \If{Node $N_{\Phi}$ still has unexpanded alternative states}
                    \State \Return EXPAND($N_{\Phi}$)
                \Else 
                    \State $N_{\Phi}$ = \Call{GetBestChild}{$N_{\Phi},C$}
                \EndIf
            \EndWhile
            \EndFunction

            \State
            \Function {Simulation}{$N_{\Phi}$}
                \State $N_{\Phi_{\rm tmp}} = N_{\Phi}$
                \State $C_{\rm count} = 0$
                \While{$C_{\rm count} < N_{\rm SimStepNum}$}
                    \State Generate the next state $\Phi_{\rm next}$ of $\Phi_{\rm tmp}$ by expert algorithm or random algorithm.
                    
                    \State $C_{\rm count} = C_{\rm count} + 1$ 
                    \State $N_{\Phi_{\rm tmp}}$ = $N_{\Phi_{\rm next}}$
                \EndWhile
                \State Calculate the evaluate value of node $N_{\Phi}$: $J_{\rm SimStepL}$,$J_{\rm StepExp}$,$J_{\rm marginExp}$,$J_{\rm disToPar}$
                \State 
                \Return the forward distance of $N_{\Phi_{\rm tmp}}$, $J_{(.)}$
            \EndFunction
            
    \algstore{myalg}
	\end{algorithmic}
\end{algorithm*}

\begin{algorithm*}[ht]
	\begin{algorithmic}[1]
	    \algrestore{myalg}
            \Function {BackUpdate}{$N_{\Phi}$, $J_{(.)}$}
                \State $score = \sum \omega_{(.)}J_{(.)}$
                \State $X(N_{\Phi}) = score$
                \State $N_{\rm vist}(N_{\Phi})$ = 1
                \While{Node $N_{\Phi}$ is not the root node}
                    \State $N_{\Phi}$ = the parant of $N_{\Phi}$
                    \State $N_{\rm vist}(N_{\Phi})$ = $N_{\rm vist}(N_{\Phi})$ + 1
                    \If{$X(N_{\Phi}) < score$}
                        \State $X(N_{\Phi}) = score$
                    \EndIf
                \EndWhile
            \EndFunction

        \State
        \Function {Expand}{$N_{\Phi}$, $S_{\rm sequence}$}
            \State Randomly select a state $\Phi_{\rm random}$ from the set of candidate states, which has not expanded, of node $N_{\Phi}$.
            
            \State Set $N_{\Phi_{\rm random}}$ to be a child node of $N_{\Phi}$
            \State 
            \Return $N_{\Phi_{\rm random}}$
        \EndFunction

        \State
        \Function {GetBestChild}{$N_{\Phi}$,$C$}
            \State
            \Return $    \underset{N_{\Phi_i} \in \ children \ of \ N_{\Phi}}{\rm arg\ max} (X(N_{\Phi_i}) + C\cdot \sqrt{\frac{2\cdot{\rm ln}N_{\rm vist}(N_{N_{\Phi}})}{N_{\rm vist}(N_{\Phi_i})}} )$
        \EndFunction
	\end{algorithmic}
\end{algorithm*}

\end{document}